# Segment Anything for Video: A Comprehensive Review of Video Object Segmentation and Tracking from Past to Future

Guoping Xu[1], Jayaram K. Udupa[2], Yajun Yu[1], Hua-Chieh Shao[1], Songlin Zhao[3], Wei Liu[3], You Zhang[1]*

[1]The Medical Artificial Intelligence and Automation (MAIA) Laboratory, Department of Radiation Oncology, University of Texas Southwestern Medical Center, Dallas, TX 75390, USA

[2]Medical Image Processing Group, Department of Radiology, University of Pennsylvania, Philadelphia, PA 19104, USA

[3]Department of Radiation Oncology, Mayo Clinic, Phoenix, 85054, AZ, USA

*Corresponding author: You.Zhang@UTSouthwestern.edu

**Abstract**

Video Object Segmentation and Tracking (VOST) presents a complex yet critical challenge in computer vision, requiring robust integration of segmentation and tracking across temporally dynamic frames. Traditional methods have struggled with domain generalization, temporal consistency, and computational efficiency. The emergence of foundation models like the Segment Anything Model (SAM) and its successor, SAM2, has introduced a paradigm shift, enabling prompt-driven segmentation with strong generalization capabilities. Building upon these advances, this survey offers a comprehensive review of SAM/SAM2-based methods for VOST, structured along three temporal dimensions: past, present, and future. We examined strategies for retaining and updating historical information (past), approaches for extracting and optimizing discriminative features from the current frame (present), and motion prediction and trajectory estimation mechanisms for anticipating object dynamics in subsequent frames (future). In doing so, we highlighted the evolution from early memory-based architectures to the streaming memory and real-time segmentation capabilities of SAM2. We also discussed recent innovations, such as motion-aware memory selection and trajectory-guided prompting, that aim to enhance both accuracy and efficiency. Finally, we identified remaining challenges—including memory redundancy, error accumulation, and prompt inefficiency—and suggested promising directions for future research. This survey provides a timely and structured overview of the field, aiming to guide researchers and practitioners in advancing the state of VOST through the lens of foundation models.

## 1. Introduction

With the growing popularity of deep learning [1-3], which enables automatic feature extraction from large-scale training data, significant advancements have been made in various computer vision tasks such as image classification [2, 4-6], object detection [7-10], object recognition [11, 12], and semantic segmentation [13-16]. However, most of these methods have predominantly targeted 2D static images, largely due to the memory and computational limitations of contemporary GPU hardware [17-19]. Recent advancements in GPU performance, driven by the rising demand for artificial intelligence and innovations in chip architecture, have begun to alleviate these constraints. Consequently, research efforts have expanded to address more complex and temporally dynamic tasks, such as video classification [20-22], object detection and tracking in video streams [23-25], analysis of 3D point clouds [26], and segmentation of anatomical structures in 3D medical volumes [27, 28].

In particular, video object segmentation and tracking (**VOST**[1]) has recently emerged as a prominent research focus in both computer vision and medical image analysis. This trend is driven by rapid technological advancements and the increasing demand for intelligent systems in applications such as mobile devices [29], video surveillance [30], robot-assisted surgery [31], and other related domains. VOST generally comprises two fundamental tasks: object segmentation and object tracking, when processing a video as a sequence of separable frames. Object segmentation involves delineating every pixel belonging to the object of interest in each frame, while object tracking aims to maintain the identity and spatial continuity of the object across successive frames. Although these sub-tasks can be considered separately, they are inherently interdependent. On one hand, accurate segmentation enhances tracking performance by providing precise and reliable object boundaries, which helps address challenges such as scale variation and occlusion. On the other hand, robust tracking facilitates more accurate segmentation by supplying consistent object localization over time, thus mitigating difficulties associated with rapid motion or the presence of similar-looking objects [32]. Hence, one of the core questions of VOST is how to *jointly* perform accurate pixel-wise segmentation and reliable identity tracking of objects over time, despite various real-world challenges, such as motion blur, shape changes, object interactions, and occlusion.

[1] VOST methods include supervised, unsupervised, semi-supervised, weakly-supervised, and interactive approaches. This study focuses on promptable VOST, where a prompt (e.g., mask, points, or box) is provided on the first frame, covering the semi-supervised, weakly-supervised, and interactive settings. We refer to this as VOST throughout the paper unless noted otherwise. Furthermore, while our method is applied to 2D images involving both rigid motion and non-rigid deformations, the framework is also applicable to volumetric object motion in dynamic tomographic imaging.

Over the past decade, numerous methods have been proposed for Video Object Segmentation and Tracking [30]. Most of these approaches follow an encoder-decoder framework, in which guidance information from preceding frames is integrated to support the segmentation of the current frame (see **Figure 1**). A representative work in this line is STM (Space-Time Memory Networks) [33], which introduces a memory mechanism that stores features from all past frames along with their object masks to guide the segmentation process. In contrast to earlier methods that rely solely on the first frame or the previous frame, STM efficiently utilizes information from multiple past frames through a dense memory storage and fusion algorithm in feature space. This abundant use of guidance information, coupled with efficient memory operations, enabled STM to achieve state-of-the-art performance on several public benchmark datasets at the time of its release. Following this paradigm, a series of subsequent works built upon the core idea of leveraging memory networks to extract and propagate historical information. These approaches aim to overcome specific limitations of STM, including the shared encoder design addressed by STCN [34], memory management improvements in XMem [23], the use of multi-scale memory features in [35], and enhanced temporal correspondence modeling in [36], among others.

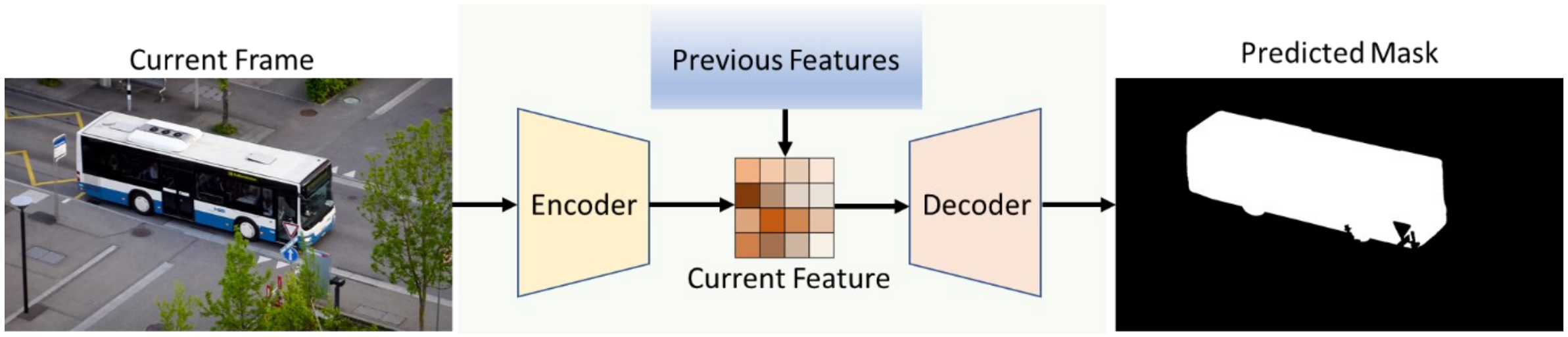

***Figure 1.*** *The pipeline of the basic architecture for VOST. The encoder extracts features from the current frame, while features from preceding frames are incorporated to update the current features by providing temporal-spatial cues. This facilitates the recognition of target objects and the differentiation of other object regions. The decoder generates the predicted mask for the current frame.*

Despite significant progress in recent years, current VOST approaches still face notable limitations that hinder their practical deployment. First, many existing methods exhibit limited generalization capability to unseen domains or new datasets, primarily due to being trained on narrowly curated or domain-specific data. Second, the optimal integration of segmentation and tracking remains an open challenge. In particular, it is unclear how to best leverage the historical temporal information to guide segmentation in the current frame and enhance tracking in subsequent frames for temporal consistency. Third, existing methods often struggle to achieve both high accuracy and computational efficiency, especially under complex or dynamic real-world scenarios. Lastly, the reliance on extensive manual annotations poses a significant barrier. Labeling every frame in a video is labor-intensive and time-consuming, making it imperative to explore more efficient training paradigms, such as semi-supervised or weakly-supervised learning [30, 32].

Inspired by the success of large language models such as the GPT (Generative Pre-trained Transformer) series [37-39], the **Segment Anything Model (SAM)** [40] was introduced in 2023 as a prompt-based foundation model for image segmentation. SAM was trained on a newly curated dataset, SA-1B**,** which contains over 1 billion masks across 11 million images. Due to its remarkable zero-shot and few-shot generalization capabilities, SAM has garnered substantial attention in the research community. It represents a significant paradigm shift—from training task-specific models to fine-tuning a powerful pre-trained foundation model through interactive prompting [35].

Capitalizing on this advancement, an increasing number of studies have begun to investigate the integration of SAM into VOST as a promising approach to address the aforementioned challenges, including limited generalization, suboptimal temporal consistency, and annotation inefficiency. One of the most widely adopted strategies involves integrating SAM with tracking modules, leveraging its strong feature representation capabilities for segmentation. For instance, TAM [41] incorporates XMem, an efficient memory-based tracking framework, into the SAM pipeline to enable interactive segmentation and tracking in videos. In this approach, SAM generates an initial object mask, which is then used by XMem to propagate the segmentation across frames by modeling temporal correspondences, thereby enhancing both tracking robustness and segmentation accuracy. Similarly, SAM-Track [42] integrates DeAOT, a video object segmentation and tracking model with an identification mechanism that associates multiple targets within a shared high-dimensional embedding space. This combination enables effective multi-object segmentation and tracking by enforcing temporal coherence. In another line of work, SAM-PT [43] proposes a point-centric tracking method that is coupled with SAM to support efficient and accurate video segmentation. In contrast to the above methods, which typically apply the tracking module after SAM, HQTrack [44] adopts a reverse architecture (See in **Figure 2**). It first employs a video multi-object segmentation model to generate coarse predictions, which are then used to automatically extract prompts for SAM. This approach enables SAM to refine its per-frame segmentation outputs, allowing the automation of the prompting process with enhanced accuracy.

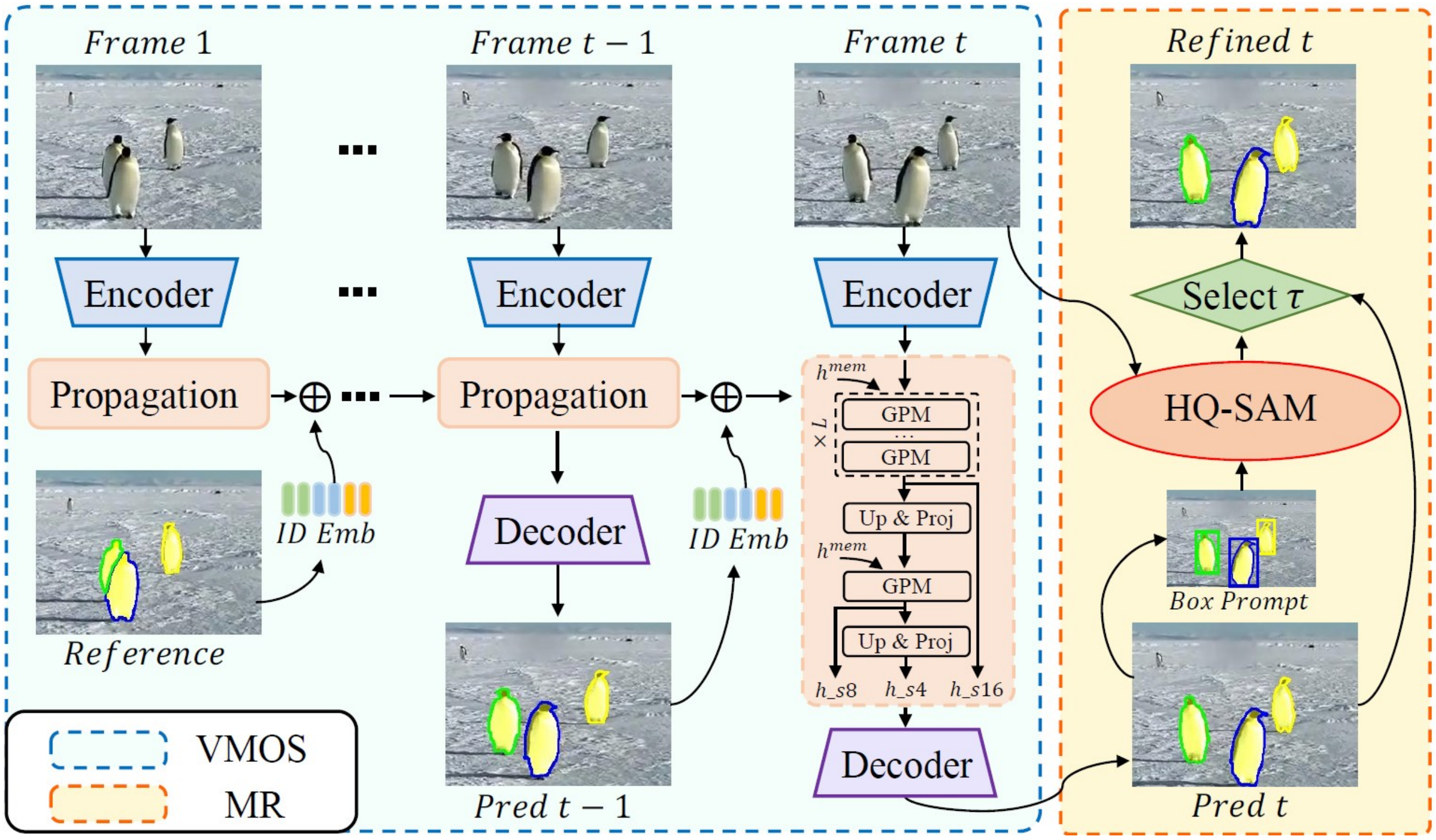

***Figure 2**. Schematic of HQTrack. The framework consists of two main components: a video multi-object segmenter (VMOS) and a SAM-based mask refiner (MR). Source: reproduced from [44].*

Although SAM-based VOST methods have demonstrated improved generalization and segmentation performance, several significant limitations remain: (1) Compatibility: The integration between the tracker and SAM may be suboptimal, as SAM is originally designed for static image segmentation and may not perform consistently on video frames. (2) Error correction: These methods often lack a mechanism for correcting errors during the tracking and segmentation process. As a result, mistakes can accumulate and propagate across frames, degrading overall performance. (3) Efficiency: Inference speed is a critical factor for real-world VOST applications, particularly in scenarios such as autonomous driving and robot-assisted surgery. However, the computational overhead introduced by SAM's heavy Transformer-based architecture significantly limits its practicality in such time-sensitive environments.

In 2024, a unified model for real-time video and image segmentation, SAM2, was introduced [45]. Building upon the same principles as its predecessor, namely, prompt-based interaction and a data generation engine, SAM2 was trained on a newly constructed large-scale dataset, SA-V, which comprises 35.5 million masks across 50.9K videos. Compared to previous approaches, SAM2 achieves higher accuracy in video segmentation and demonstrates a 6× improvement in speed over the original SAM model for image segmentation. This raises two key questions: *Have all challenges in video object segmentation and* *tracking (VOST) been solved by SAM2? If not, what progress has been made in SAM2-based VOST, and what are the remaining challenges for future research?*

To address these questions, we present a systematic review of the existing literature on VOST methods, mainly focusing on SAM/SAM2-based methods. We categorize these approaches into three conceptual stages—past, present, and future—focusing respectively on (1) how historical information from previous frames is stored and updated (past), (2) how discriminative features are effectively and efficiently learned for the current frame (present), and (3) how object trajectories are accurately estimated and tracked in subsequent frames (future). For each category, we analyze the strengths and limitations of representative methods and offer insights into promising directions for future research in VOST.

Although several surveys have systematically reviewed the progress in video object segmentation and tracking (VOST) [30, 32, 46, 47], they largely predate the emergence of foundation models like SAM2 and therefore do not reflect these recent advancements. More recent works have begun to examine SAM2's performance in specific applications, for example, segmentations of camouflaged objects [48] and biomedical images and videos [49]. Additionally, surveys such as [50] and [51] provide a systematic overview of SAM- and SAM2-based methods in VOST. These surveys provide valuable historical context, introduce foundational concepts, and highlight representative works, collectively offering a solid foundation for understanding and keeping pace with ongoing developments in this rapidly evolving field.

In contrast to previous works, our review provides a focused exploration of the evolution from earlier approaches to SAM2-based methods for Video Object Segmentation and Tracking (VOST). We examined this progression through the lens of three fundamental components: memory update, feature learning, and motion tracking. These components are conceptually aligned with the temporal dimensions of VOST processing—past, present, and future, respectively. We traced the development of SAM2 from its initial implementations to its most recent innovations, highlighting how contemporary methods have leveraged SAM2 to address long-standing challenges in VOST. Additionally, we critically evaluated the strengths and limitations of these approaches and suggested promising directions for future research. An overview of this survey is presented in **Figure 3**, where we summarized representative methods and structured our discussions.

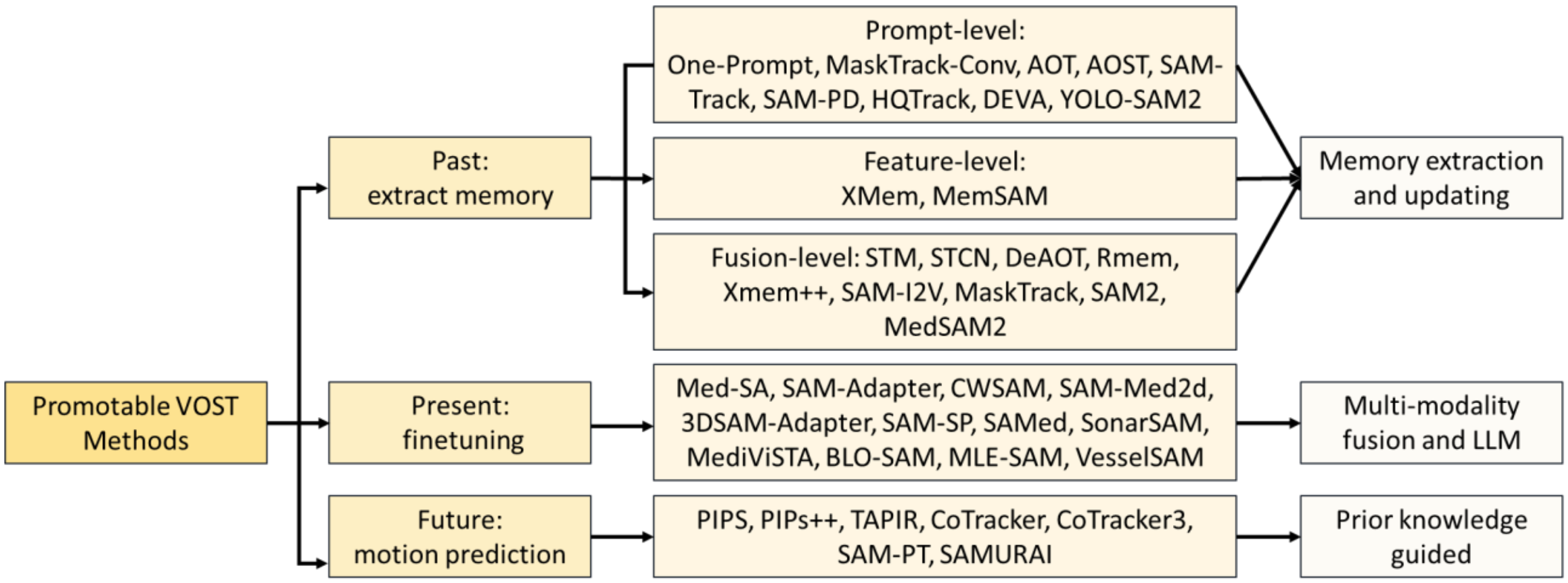

***Figure 3**. An overview of the promptable VOST methods discussed in this survey. We categorized existing approaches into three groups based on their focus on addressing the VOST task: (1) extracting and storing memory from previous frames; (2) fine-tuning SAM/SAM2-based methods to learn representative features for the current frame; and (3) modeling object motion or trajectory for future frames. On the right side, we also highlighted future directions, including memory updating strategies, multi-modality fusion, and prior-knowledge-guided motion prediction.*

In summary, our main contributions in this survey are as follows:

- We provided a comprehensive and balanced review of VOST development from past to present, focusing on the evolution and connection from previous methods to SAM2-based approaches.
- We reviewed and categorized recent SAM2-based methods for VOST, with a particular focus on the three core components: memory update (past), feature learning (present), and motion tracking (future).
- We provided a comprehensive overview of benchmark datasets and evaluation metrics commonly used in VOST research.
- We identified key challenges and emerging trends, offering insights into potential future research directions in the field.

## 2. Prerequisite: SAM and SAM2

The Segment Anything Model (SAM) is a pioneering framework designed for 2D image segmentation guided by various forms of prompts, including points, bounding boxes, and masks. It represents the first promptable, general-purpose segmentation foundation model, trained on a large-scale natural image dataset comprising over 1 billion masks and 11 million images. SAM is built upon a transformer-based architecture and consists of three main components: an image encoder, a prompt encoder, and a mask decoder (see **Figure 4**). The image encoder is tasked with extracting rich visual features from high-resolution inputs and is pre-trained using the Masked Autoencoder (MAE) self-supervised learning strategy [52], providing a strong initialization for handling complex segmentation tasks. The prompt encoder captures the spatial context of the target objects. For sparse prompts such as points and bounding boxes, it generates embeddings by summing learned prompt embeddings with positional encodings. For dense prompts like masks, it applies a lightweight convolutional network to derive prompt features, which are then added to the image embeddings before being passed to the mask decoder. This design enables SAM to flexibly integrate diverse prompt types and produce accurate segmentation results across a wide range of input conditions.

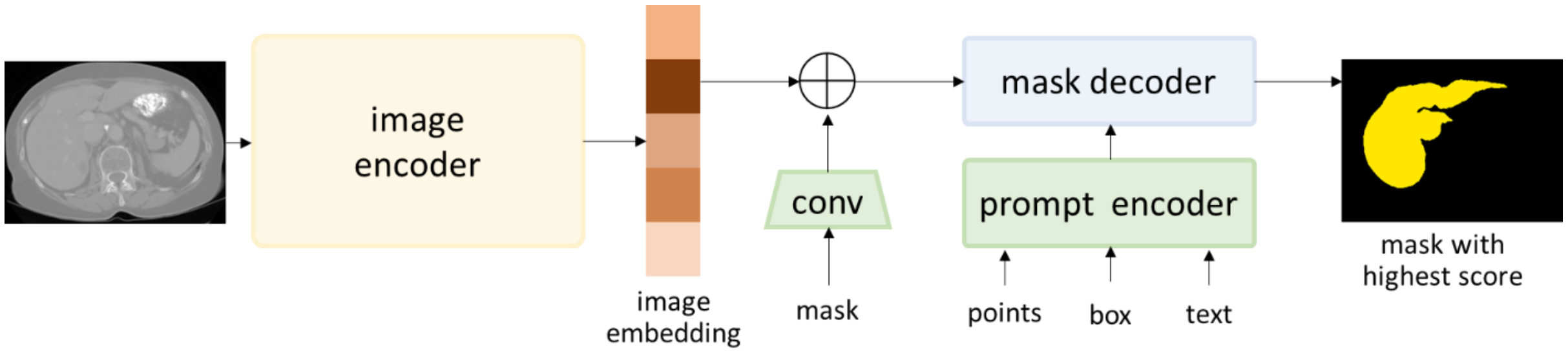

***Figure 4***. *A schematic of SAM. The architecture primarily consists of a heavyweight image encoder, a prompt encoder, and a mask decoder, which serve to extract features, recognize objects, and generate segmentation masks.*

The image and prompt embeddings are then passed into a lightweight mask decoder, which is responsible for generating the final segmentation masks. Within the decoder, prompt self-attention and cross-attention mechanisms are employed to update and fuse features from both the prompt and image embeddings. The resulting fused representations are subsequently upsampled to produce high-resolution segmentation outputs. This design enables SAM to perform segmentation tasks efficiently and accurately in a prompt-driven manner, making it highly adaptable across diverse application scenarios.

Building upon the core principles of SAM, SAM2 further improves segmentation performance and enhances flexibility for both image and video inputs. Its overall architecture is illustrated in **Figure 5** and comprises six key components: an image encoder, a prompt encoder, a mask decoder, memory attention, a memory encoder, and a memory bank.

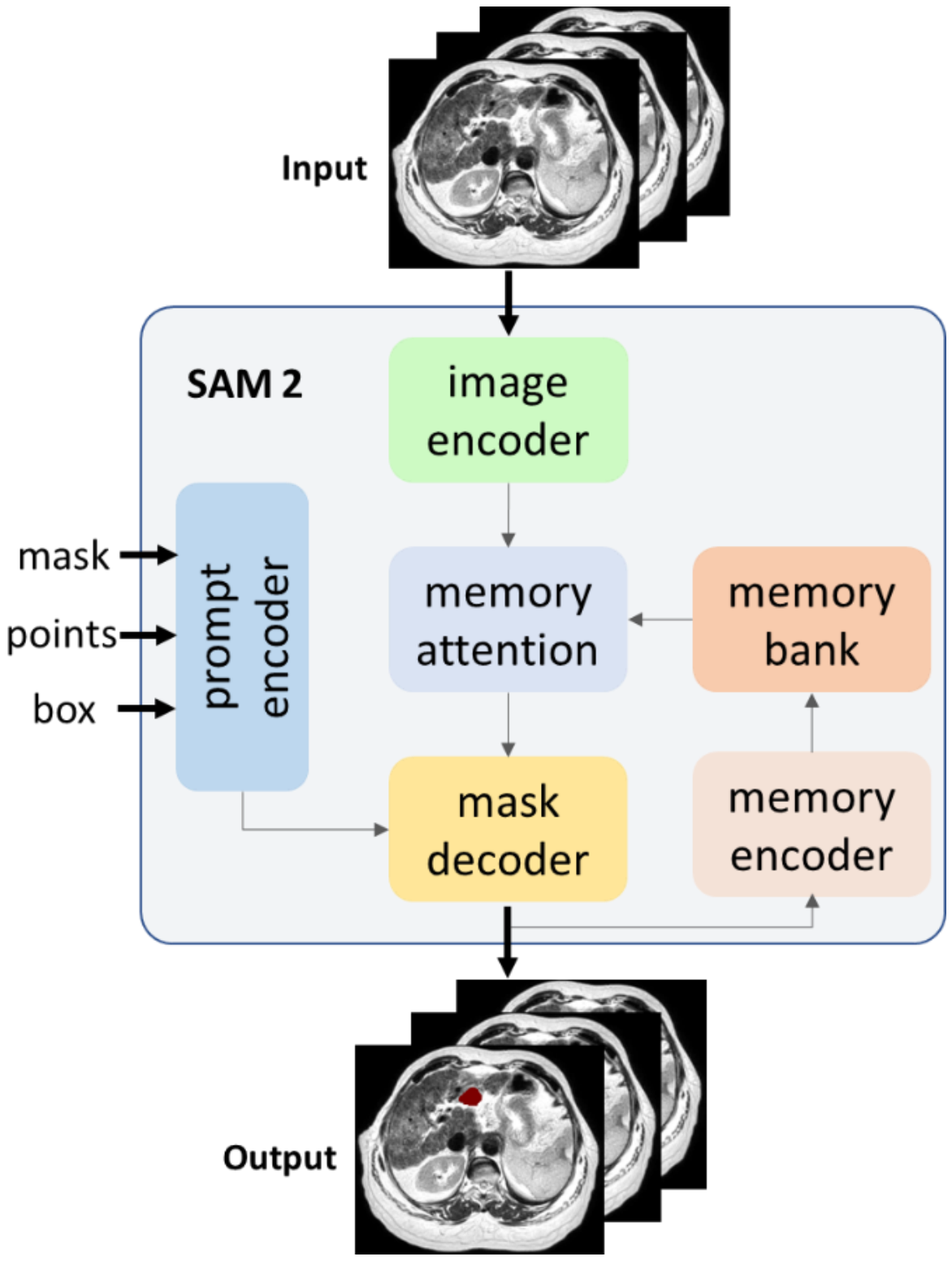

*Figure 5. The SAM2 architecture. A memory bank stores features from previous frames, while the memory attention module updates the current frame's features based on the stored memory. The memory encoder then fuses features from the image and the predicted mask for future use.*

In contrast to the original SAM, SAM2 introduces a streaming memory mechanism designed to encode, update, and store information from previous frames, thereby facilitating temporal consistency and improving segmentation accuracy over time. Specifically, the memory attention module conditions the current frame's features using both the historical memory—comprising past frame features and prediction masks—and any new prompts. This module contains $L$ transformer blocks that employ self-attention to refine the current frame features and cross-attention to integrate historical information, enabling the model to localize target objects more effectively.

The memory encoder fuses features from the image encoder and the predicted mask using a lightweight convolutional architecture, while the memory bank stores this fused information for future reference. A first-in and first-out (FIFO) queue strategy is employed to retain a fixed number of the most recent frames, balancing memory capacity and computational efficiency.

To support real-time performance and maintain high segmentation accuracy, SAM2 adopts Hiera [53]—a multi-scale hierarchical vision transformer pretrained using Masked Autoencoders (MAE)—as its image encoder. This allows SAM2 to extract rich, multi-scale representations from high-resolution inputs, further boosting its segmentation capabilities across varied scenarios.

In summary, SAM has laid a promising foundation for prompt-based image segmentation, establishing a new paradigm for general-purpose segmentation models. Its successor, SAM2, extends this capability to video object segmentation and tracking (VOST), offering improved inference speed while maintaining competitive segmentation accuracy. Building upon these two foundational models, this review provides a comprehensive overview of recent advances in VOST. In the following sections, we delve into three key aspects of VOST: (i) how *past* information (memory) is retained and retrieved, (ii) how *current* frame features are extracted and updated based on stored memory, and (iii) how efficient object tracking is performed to support segmentation in *future* frames.

## 3. Past: how to memorize and update the historical features

Effectively capturing and utilizing historical information is crucial for accurate VOST. This challenge centers around two fundamental questions: (i) how to select and retain meaningful features from previous frames, and (ii) how to dynamically update and adapt these features to support current frame segmentation and future-frame tracking. To systematically address the first question, we categorize existing approaches into three levels based on how historical information is represented and retained: prompt-level, feature-level, and fusion-level. Building on this framework, we further explore strategies relevant to the second question, focusing on pruning-based and time-scale-based methods employed in SAM2-based VOST models for dynamic memory bank updates. In the following subsections, we present a structured review of representative techniques under each category, highlighting their design choices, strengths, and limitations.

*(1) Prompt-level: Methods that encode temporal memory mainly through prompt tokens or prompts derived from historical frames, guiding segmentation with temporal context.*

Various types of prompts—such as points, scribbles, bounding boxes, masks, and text descriptions—can be applied in the initial frames of a video to initiate VOST. These prompts provide essential semantic and spatial cues that enhance the accuracy of object localization and segmentation across frames. A central challenge in VOST is the effective encoding of temporal memory from sequential prompts or prompt-derived tokens to address issues such as object deformation, scale and position changes, motion blur, occlusions, and background ambiguity.

A recently proposed foundation model, One-Prompt [54], designed for 3D medical volume segmentation, offers inspiring insights for tackling this challenge for VOST. The architecture of One-Prompt, illustrated in **Figure 6(a)**, includes a key component called the One-Prompt Former module (see **Figure 6(b)**). At the heart of this module is the Prompt Parser (**Figure 6(c)**), which integrates both prompt and image embeddings from a template frame with features from the current frame (referred to as the query image). Specifically, positional information is encoded into the prompt memory to preserve spatial context and is added to learnable embeddings. These enhanced prompt representations, along with template features, are then used in a cross-attention mechanism to guide segmentation in the query frame. In this framework, prompt memory provides spatial guidance, helping to localize foreground objects and suppress background interference.

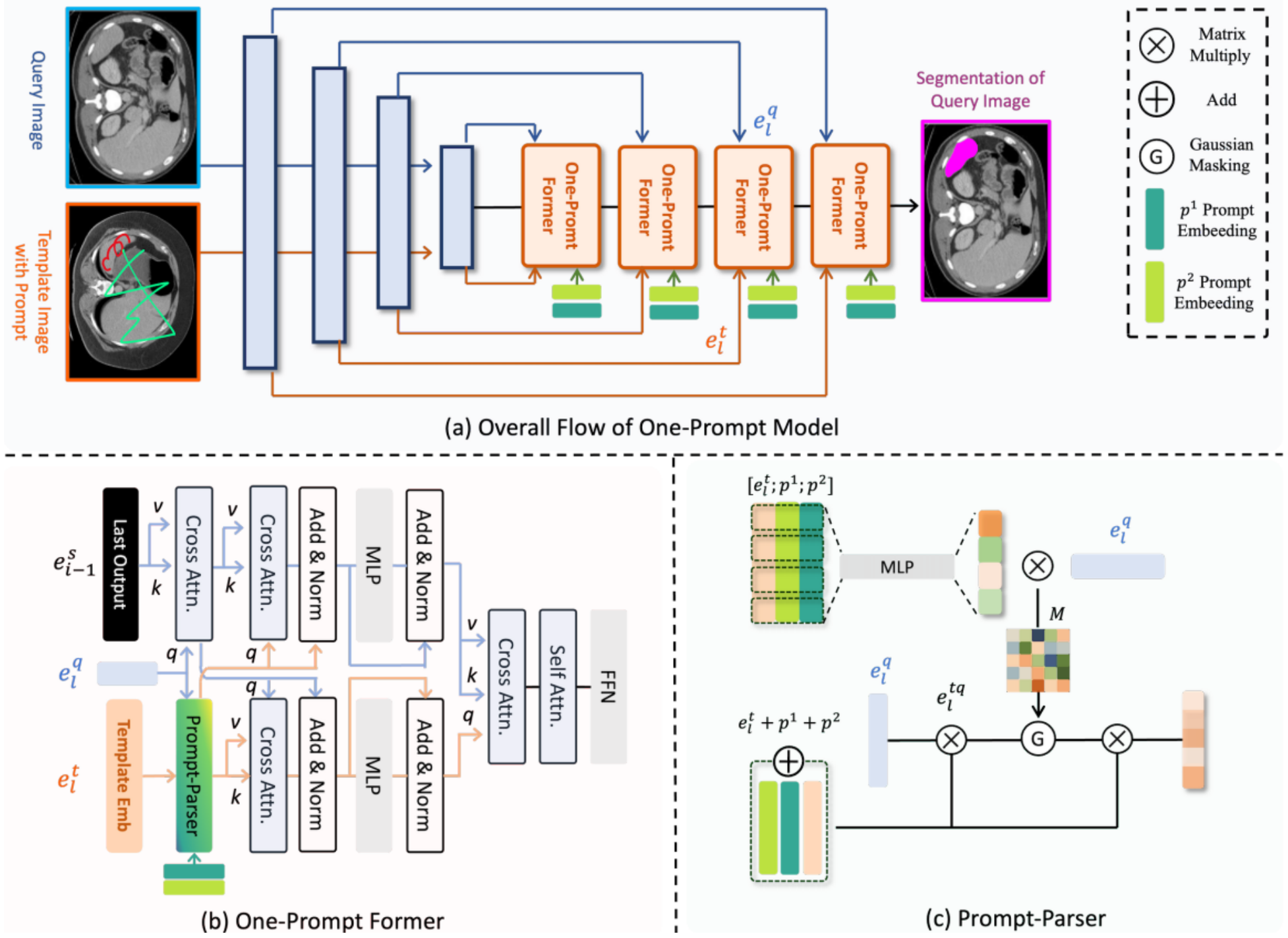

*__Figure 6.__ Illustration of One-Prompt [54]. (a) The architecture of the One-Prompt model; (b) the proposed One-Prompt Former module; and (c) the design of the Prompt-Parser module. Reproduced from [54], with permission from IEEE.*

Like the idea of One-Prompt—where prompts provide a rough indication of the target's location and region—a range of prompt (mask) propagation-based methods have been developed for VOST to further incorporate temporal continuity [30, 32]. Unlike One-Prompt, which is limited to static image segmentation, these VOST methods leverage sequential prompts across frames to maintain consistency over time and improve segmentation accuracy in dynamic scenes. In MaskTrack ConvNet [55], the predicted mask from the previous frame is used as an additional input channel to guide the segmentation of the current frame. To integrate motion information, a variant of MaskTrack introduces an optical flow-based model—EpicFlow [56]—augmented with Flow Fields matching [57] and convolutional boundary refinement [58], generating an optical flow magnitude field as the temporal prompt. However, this strategy heavily depends on the accuracy of the predicted masks and the quality of the optical flow estimation. It is particularly vulnerable in cases of large inter-frame motion, where misalignment of the prompt may occur, leading to suboptimal segmentation performance. Moreover, this method relies solely on a single previous prediction as the prompt, thereby underutilizing rich historical information that could enhance performance.

To address these limitations, recent works such as AOT [59] and AOST[60], introduce an Identification (ID) mechanism for encoding multi-object mask embeddings, which are propagated across frames to guide subsequent predictions (see **Figure 7**). Coupled with a Long Short-Term Transformer (LSTT) module, these frameworks can retain and utilize long-term object-specific memory, thereby enabling more robust tracking and segmentation in dynamic video scenarios. In addition, other techniques have been introduced to further improve the fidelity of propagated masks. These include contour evolution strategies [61], bidirectional propagation [62], optical flow [63], and reinforcement learning [64], all of which aim to encourage the model to focus more precisely on plausible regions informed by historical masks.

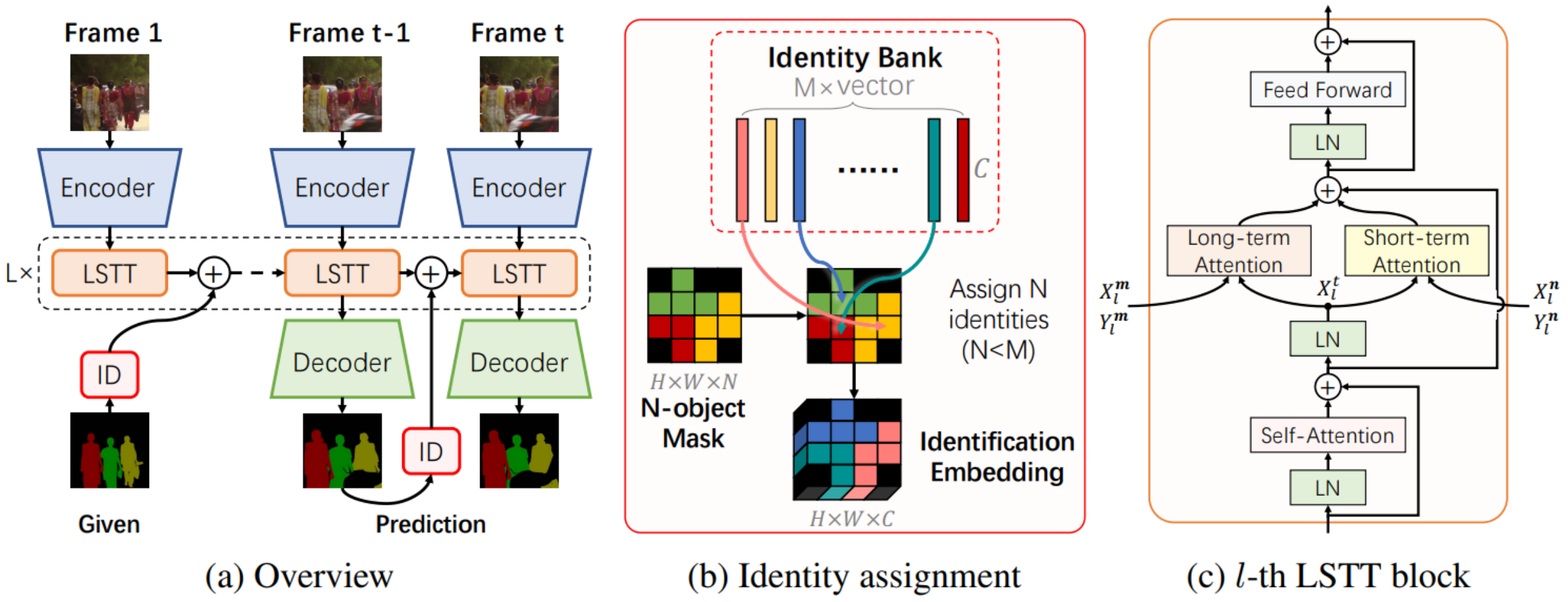

***Figure 7**. (a) The pipeline of Associating Objects with Transformers (AOT); (b) an illustration of identity assignment for embedding ID information into features from previous frames; and (c) the structure of an LSTT block, designed for long-term and short-term memory extraction. Reproduced with permission from [59].*

Despite incorporating various techniques to enhance mask propagation, many task-specific approaches in VOST lack flexibility when handling diverse prompt types, such as points and bounding boxes.

Furthermore, the sub-optimal quality of propagated masks and limited generalization ability hinder their overall performance and practical deployment. SAM, trained on a large-scale dataset, demonstrates a strong ability to generate high-quality masks from various prompt types and shows impressive zero-shot generalization. However, directly applying the image-based SAM to VOST yields suboptimal results due to its failure to account for temporal coherence across video frames. To address this, recent studies have explored integrating SAM with prompt-aware temporal modeling methods to better adapt it for VOST tasks.

In SAM-Track [42], SAM is employed to generate segmentation masks in conjunction with Grounding-DINO [65], enabling support for text-based prompts. These initial masks are then passed to DeAOT, which performs refined segmentation and uses the results as reference frames for subsequent predictions within the VOST framework. Similarly, FlowP-SAM [66] incorporates optical flow as an auxiliary prompt to guide SAM's frame-level segmentation. To ensure temporal consistency of object identities across frames, a sequence-level mask association module is introduced as a post-processing step, operating on a series of previously predicted masks. In SAM-PD [67], the bounding box extracted from the predicted mask of the preceding frame is propagated to the next frame, leveraging SAM's robustness to noisy or imprecise prompts. To further enhance both prompt quality and final segmentation performance, two refinement strategies are employed: multi-prompt and point-based mask refinement (see **Figure 8**). The first strategy constructs a coarse mask using multiple bounding boxes derived from the previous prediction, while the second strategy samples points from this coarse mask and uses them as new prompts to guide more accurate segmentation. Similarly, a series of studies aim to enhance SAM by automatically generating suitable prompts from previous frames to guide inference in the current frame, such as HQTrack [44], and DEVA [68]. In addition, certain SAM-based methods incorporate external object detectors to automatically generate bounding boxes, which are then used by SAM to produce frame-by-frame segmentation predictions [69, 70].

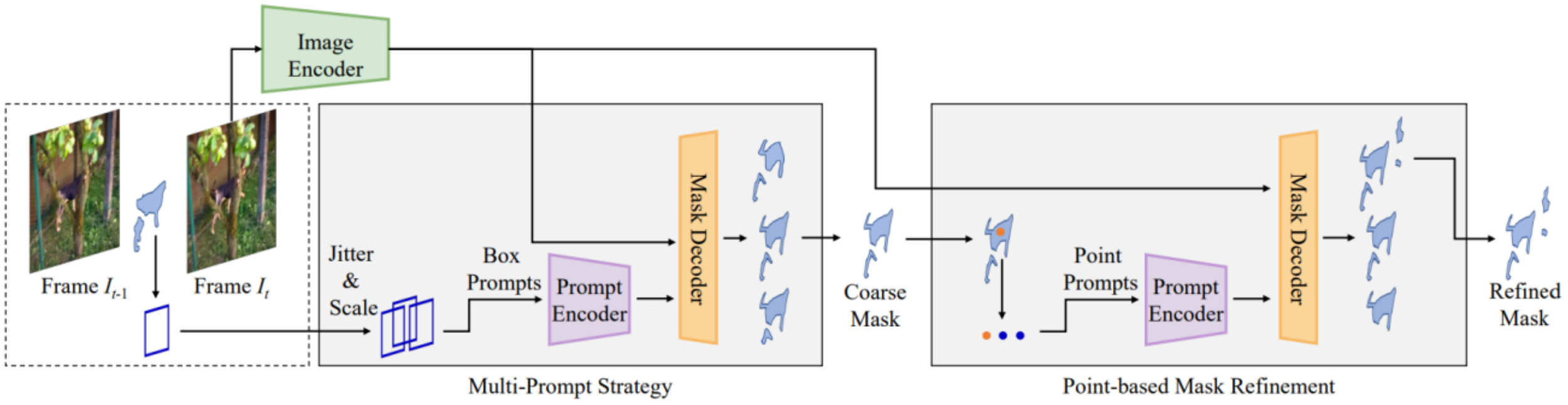

***Figure 8**. The pipeline of SAM-PD. A multi-prompt strategy generates multiple coarse masks, followed by a point-based refinement stage that selects positive and negative points to improve the final predicted mask. Source: reproduced from [67].*

However, these prompt-memory propagation-based approaches built upon SAM are inefficient, as they rely on additional prompt trackers to generate prompts and struggle to establish strong temporal consistency between successive frames, thereby hindering both inference speed and accuracy.

*(2) Feature-level: Approaches that maintain and update intermediate feature representations across frames, enabling consistent object representation over time.*

In feature-level memory-based VOST, a critical challenge lies in how to effectively store (write) past information and retrieve (read) relevant features for accurate segmentation of the current frame. XMem [23] addresses this by drawing inspiration from the Atkinson-Shiffrin memory model [71], categorizing historical features into three memory types: a rapidly updated sensory memory, a high-resolution working memory, and a compact long-term memory—each capturing different temporal scales. To read from memory efficiently and prevent memory overflow or performance degradation, XMem introduces a prototype selection and a memory potential algorithm that selectively consolidate working memory into long-term memory (as shown in **Figure 9**). Leveraging a hierarchical time-scale memory and a handcrafted memory management strategy to integrate current features with historical information, XMem achieved state-of-the-art performance on the Long-time Video datasets at the time of its release.

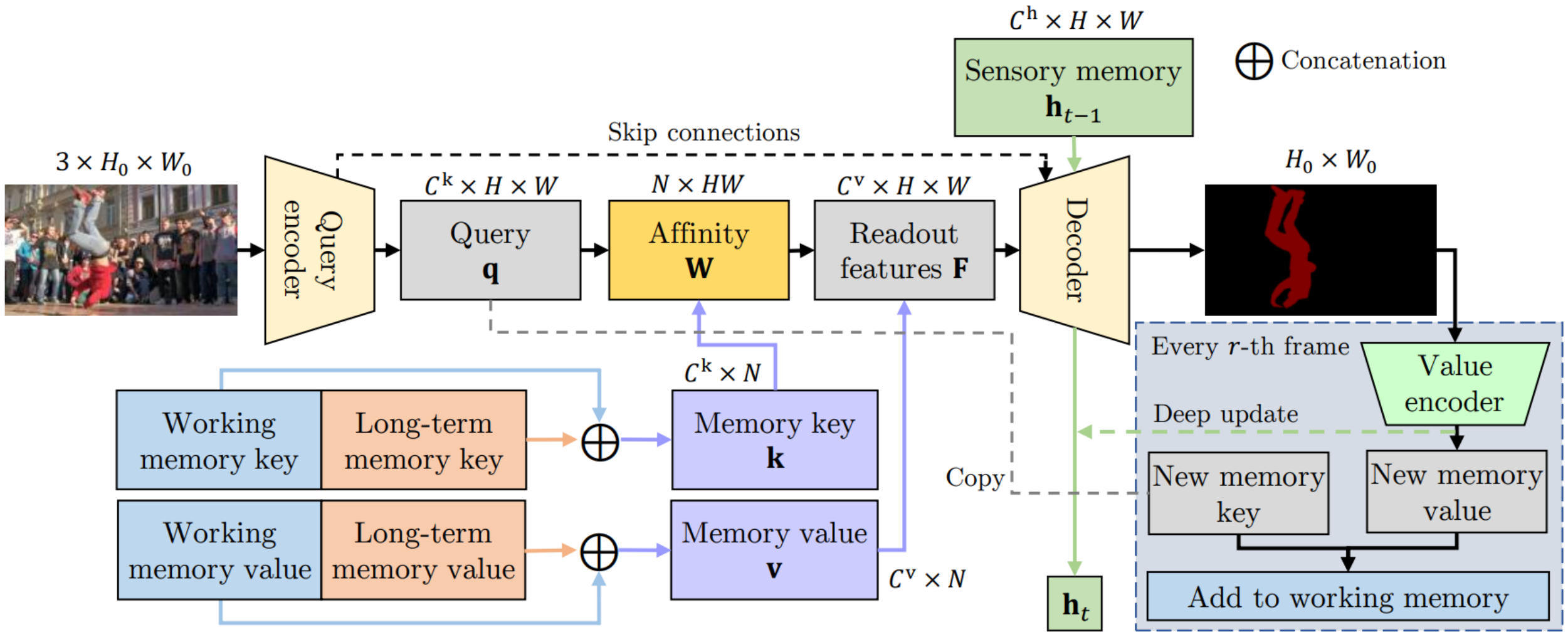

***Figure 9***. *Process of memory writing and reading of XMem. In the memory writing phase, features from previous frames and mask encoding are encoded into working memory and long-term memory. During memory reading, these stored memories are transformed into keys and values, which are used to update the features of the current query frame. Reproduced from[23], with permission from Springer.*

Building on the success of SAM-based segmentation, several subsequent methods [41, 72, 73] have integrated SAM with XMem or similar memory management modules to enhance performance in VOST, where XMem primarily serves as a prompt generation mechanism. In particular, MemSAM [73] introduces a space-time memory module that captures both spatial and temporal cues to guide the segmentation of the current frame (see **Figure 10**). Specifically, MemSAM feeds the current segmentation results into a Memory Reinforcement module designed to suppress the accumulation and propagation of noisy features while enhancing the discriminability of feature representations stored in memory. In parallel, a Sensory Memory—updated via a Gated Recurrent Unit (GRU) [70]—is used to rapidly incorporate current prompt information. The refined prompt memory is then passed through a memory encoder to generate prompt embeddings. These embeddings are subsequently organized into a Working Memory and a Long-Term Memory, which persist and evolve across frames. During inference, the model reads from the sensory, working, and long-term memory modules to guide segmentation and tracking in the next frame, promoting continuity and robustness in dynamic video scenarios.

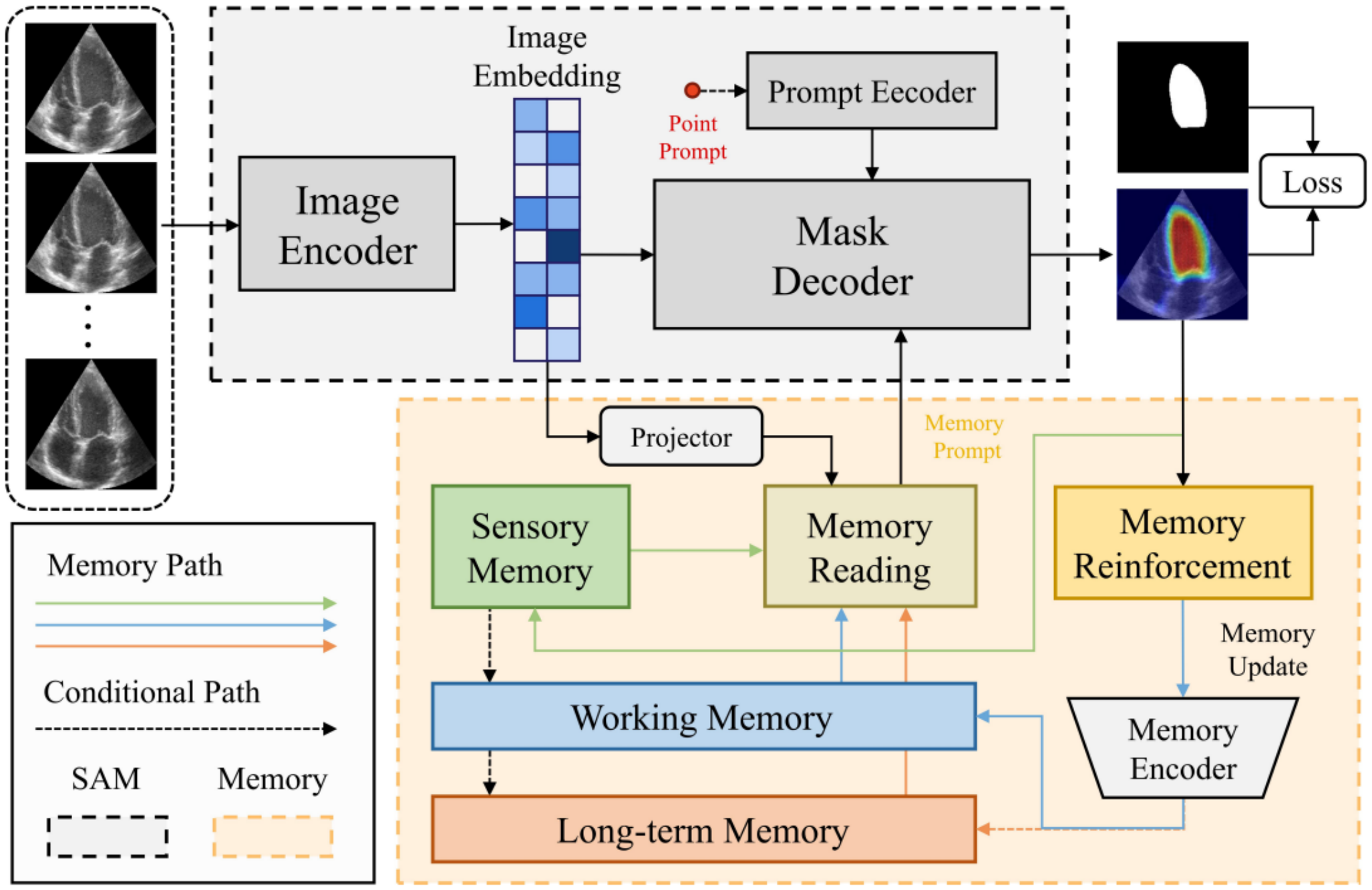

***Figure 10**. Architecture of MemSAM [73]. MemSAM primarily consists of SAM components along with dedicated memory modules. The memory module is designed to read features from the previous sensory memory, working memory, and long-term memory, each of which is encoded by its respective memory encoder. Reproduced from [73], with permission from IEEE.*

Leveraging SAM's strong capability in image feature extraction, combined with advanced memory management modules that process multiple types of memory features across different temporal scales, these approaches have achieved notable progress in Video Object Segmentation and Tracking (VOST). However, they continue to inherit critical limitations from XMem, particularly in effectively addressing core challenges such as domain generalization, fast object motion, and occlusions. These issues remain open problems and highlight the need for more adaptive and robust memory mechanisms tailored to the dynamic nature of real-world video data.

These feature-level memories updating mechanisms primarily aim to retain the most informative features while filtering out irrelevant ones, thereby improving both efficiency and segmentation performance. However, the challenge of dynamically updating the memory and selecting an optimal set of features for each current frame remains an open research question. Moreover, current approaches typically incorporate previous mask information directly into the corresponding image features, without fully modeling the relationship between image content and predicted masks. This limitation motivates further exploration, which will be discussed in the following subsection.

*(3) Fusion-level: Techniques that integrate prompt- and feature-level cues, often through multi-modal fusion modules or attention mechanisms, to enhance the robustness of temporal modeling.*

Fusion-level memory management methods in VOST focus on integrating prompt-level and feature-level cues—often through attention mechanisms or multi-modal fusion modules—to improve the robustness of temporal modeling. Earlier works such as STM [33] and STCN [34] incorporate previous mask information directly, while others like DeAOT [74], RMem [75], and XMem++ [76] handle image features and prompts separately, fusing them in a post-processing stage. However, these approaches often suffer from suboptimal feature representations, which limit the effectiveness of fusion between image and prompt features and ultimately impair the segmentation performance.

Recently, a series of works have explored the integration of SAM with memory management modules to fuse features from both image and prompt sources for updating current frame representations in object tracking and segmentation [77, 78] [79]. For example, SAM-I2V [77] (see **Figure 11**) utilizes the original SAM to extract a sequence of image features from video frames, which are then enriched with temporal context through a temporal feature integrator**.** These temporally aware features are subsequently processed by a memory selective associator**,** which manages and associates historical information from both previous image features and predicted masks. Following this, a memory prompt generator is employed to refine object-level prompts based on the selected historical memory, thereby enhancing temporal consistency and improving segmentation performance throughout the video sequence.

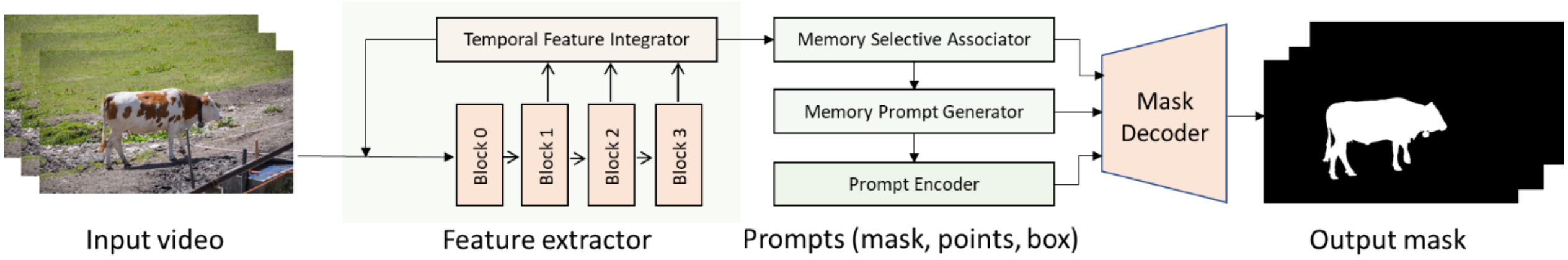

***Figure 11****. Overview of SAM-I2V [77]. The temporal feature integrator is designed to aggregate time-sequence image features, while the memory selective associator fuses image features with prediction mask features to enhance segmentation performance. Source: adapted from [75].*

In MaskTrack [79], SAM is employed to extract both initial image features and mask features. These embeddings are subsequently stored as pixel-level fused features and instance-level fused features. Notably, two specialized modules (see **Figure 12**)—the Pixel Context Transformer and the Instance Identity Transformer—are introduced to generate robust instance-level representations, aiming to maintain object identity over time for effective long-term mask propagation and accurate segmentation in VOST.

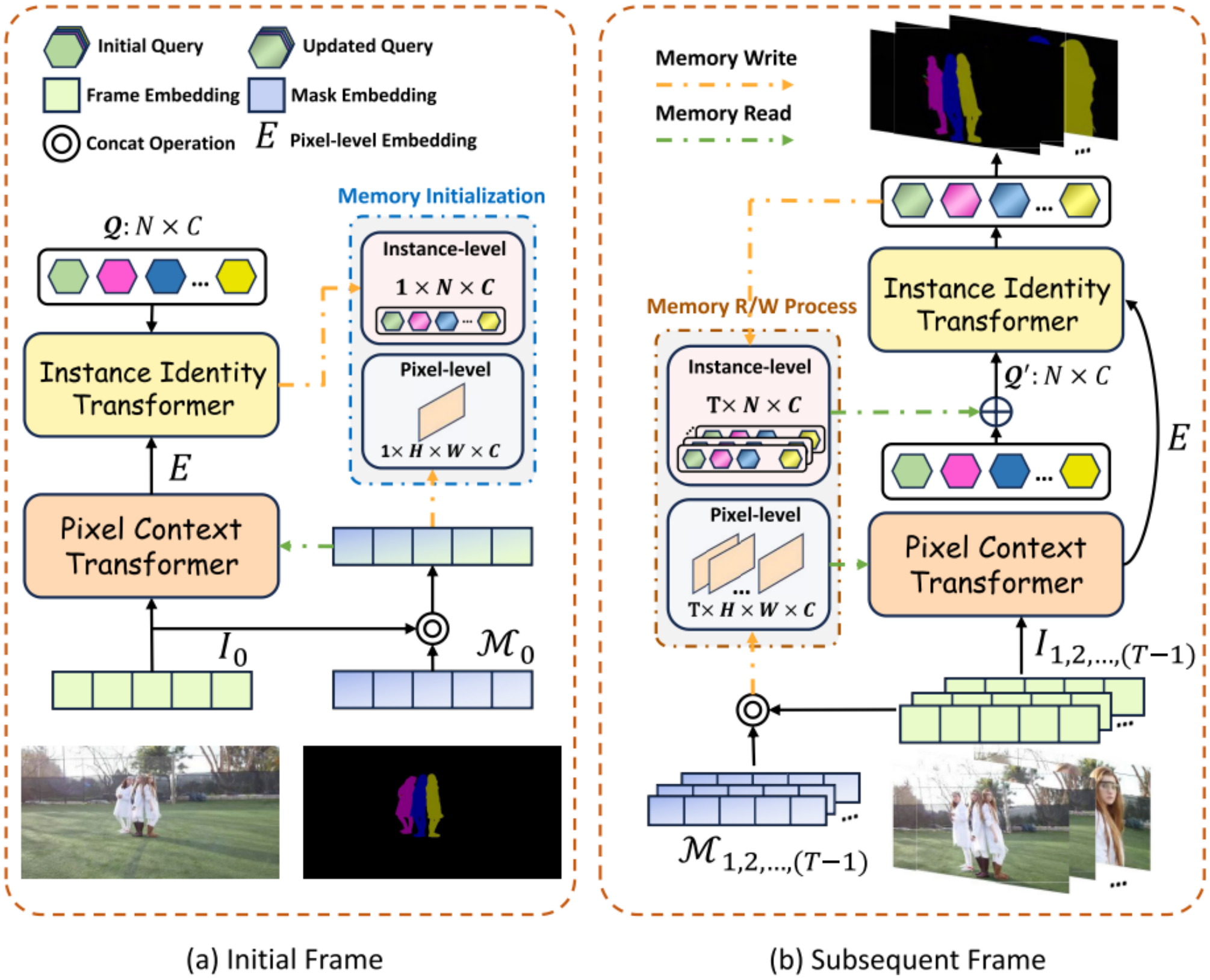

*Figure 12. Overview of MaskTrack. (a) Memory initialization for the first frame; (2) Memory reading for the subsequent frames. Source: Reproduced from [79], with permission from IEEE.*

Most SAM-based approaches incorporate memory modules with attention mechanisms to fuse image and prompt features, which are then stored as historical information for guiding future frames. However, these memory modules may not generalize well across all objects, and SAM itself may exhibit degraded performance when applied to video frames [40]. Additionally, the externally attached memory modules often lack tight integration with the original SAM architecture, leading to inefficiencies in both computation and inference speed. To address these issues, SAM2 extends the original SAM by introducing a streaming memory mechanism tailored for real-time VOST. This streaming memory consists of three core components: a memory encoder, a memory bank, and a memory attention module. Together, they facilitate (1) the fusion of image and mask features from previous frames, (2) the storage of these fused features over time, and (3) the updating of current frame features using the previously stored memory. This tightly coupled design improves both efficiency and accuracy in VOST tasks for natural scenario videos.

To enhance the generalization capability of SAM2 for medical imaging tasks, MedSAM2 was developed while retaining the original streaming memory mechanism for fusing, storing, and reading features from previous frames. Notably, MedSAM2 was fine-tuned on a large-scale dataset containing over 455,000 3D image-mask pairs and 76,000 video frames [80]. Compared to the original SAM2, MedSAM2 demonstrates significantly improved consistency and reliability in segmentation performance across diverse medical image and video datasets. This highlights its strengthened adaptability to domain-specific challenges commonly encountered in clinical imaging environments. Similarly, BioSAM2, another variant fine-tuned on biomedical images and videos, further demonstrates that domain-specific adaptation can consistently boost the segmentation performance of SAM2 [81].

*(4) Update memory efficiently for SAM2*

As previously discussed, various strategies have been proposed for extracting and storing memory representations of prompt and image features from preceding frames. In the original SAM2, features from a fixed number of past frames are stored in a memory bank using a first-in-first-out (FIFO) updating mechanism. These stored features are then uniformly passed into a memory attention module to condition the features of the current frame. However, this design overlooks the redundancy among stored frame features—particularly as the number of stored frames increases—leading to substantial computational overhead. More critically, the inclusion of erroneous or incomplete features from prior frames (e.g., stemming from inaccurate image features or segmentation masks) may result in error accumulation, thereby degrading overall segmentation performance. Furthermore, the combination of a FIFO-based updating scheme and a fixed memory bank size limits SAM2's capacity for long-term video tracking, particularly in dynamic environments with rapid scene changes, fast-moving, or self-occluding objects, where essential contextual information may be prematurely discarded [48, 78, 79].

To address these limitations, two primary strategies have been proposed for dynamically updating previous frame features stored in the SAM2 memory bank: pruning-based and time-scale-based approaches [31, 82-84]. Medical SAM2 [31] introduces a novel self-sorting memory bank that dynamically selects informative features based on confidence and dissimilarity. This design aims to improve memory quality and reduce error propagation in sequential image processing. Beyond its use in 3D medical volumes, Medical SAM2 also enables segmenting similar 2D images by treating a series of 2D slices as a video and conditioning them on a single prompt, thereby extending its utility to weakly-supervised or few-shot segmentation settings. Building on the core principles of Medical SAM2, SurgSAM2 [82] proposes an efficient frame pruning strategy that dynamically discards redundant informative features from the memory bank based on cosine similarity, retaining only the most relevant representations (see **Figure 13**).

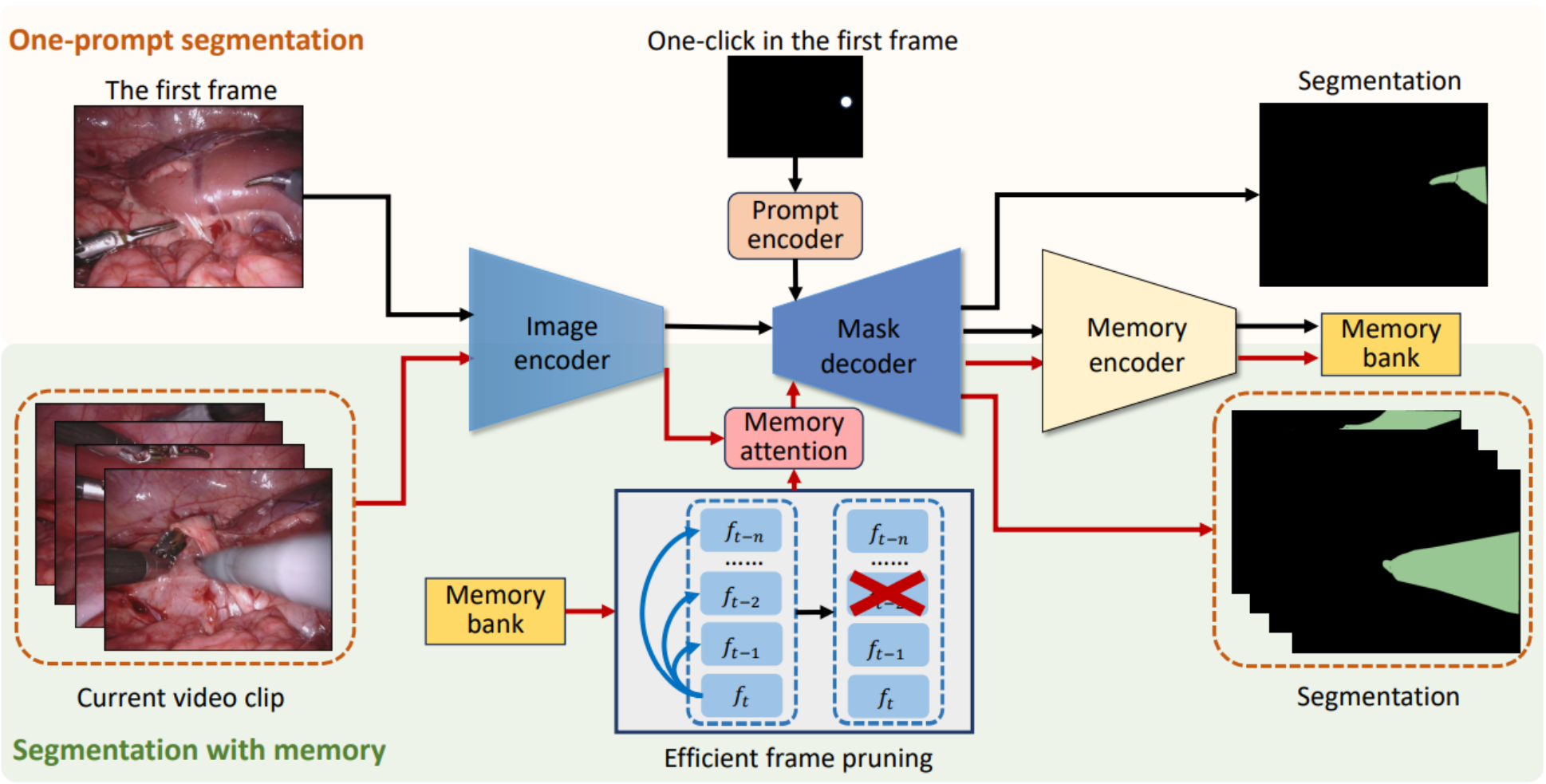

***Figure 13****. Architecture of SurgSAM2. SurgSAM2 introduces a frame-pruning strategy based on cosine similarity to discard redundant memory frames, thereby improving memory efficiency and focusing on informative temporal features. Source: reproduced from [82].*

Similarly, SAMURAI [83] extends the concept of memory pruning by introducing a motion-aware memory selection mechanism, which enhances object motion prediction and mask refinement—ultimately demonstrating robust zero-shot tracking performance (see **Figure 14**). Specifically, it incorporates a Kalman Filter (KF)-based motion model for visual object tracking to address association ambiguities, improving predictions of bounding box positions and dimensions. To guide memory selection, SAMURAI computes three key scores—mask affinity, object occurrence, and motion score—each compared against pre-defined thresholds to identify the most relevant features from previous frames. These selected memory features are then passed through a memory attention layer to update the current frame's features. This combination of motion modeling and selective memory retrieval enables accurate and efficient tracking, delivering strong zero-shot segmentation results across diverse benchmarks without requiring task-specific fine-tuning.

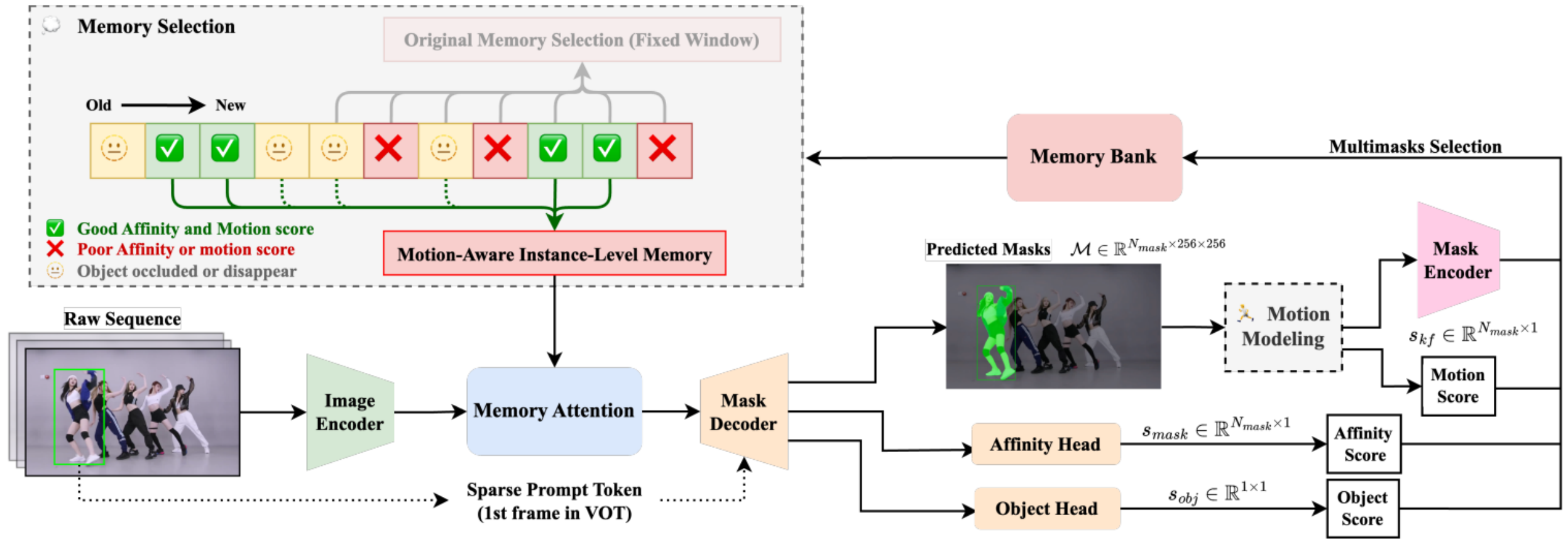

*Figure 14*. *Overview of the SAMURAI visual object tracker. The SAMURAI tracker introduces a motion-aware memory selection mechanism that predicts object motion and refines mask selection for improved tracking accuracy. Source: reproduced from [83].*

To address the error accumulation problem inherent in SAM2's fixed FIFO memory design, SAM2Long [84] introduces a training-free constrained memory tree that selectively retains more confident segmentation masks and their corresponding image features in the memory bank. This memory tree structure maintains multiple memory pathways, each comprising its own memory bank and cumulative quality score. For each input frame, the mask decoder generates three mask candidates, each conditioned on a different set of previously stored features from its memory bank. These candidates are evaluated, and the one with the highest updated cumulative score is propagated to the next time step. For example, if two memory branches are maintained, the decoder will produce a total of six candidate masks (three per branch), and the top candidate—based on its cumulative score—is selected to update the memory (See **Figure 15**).

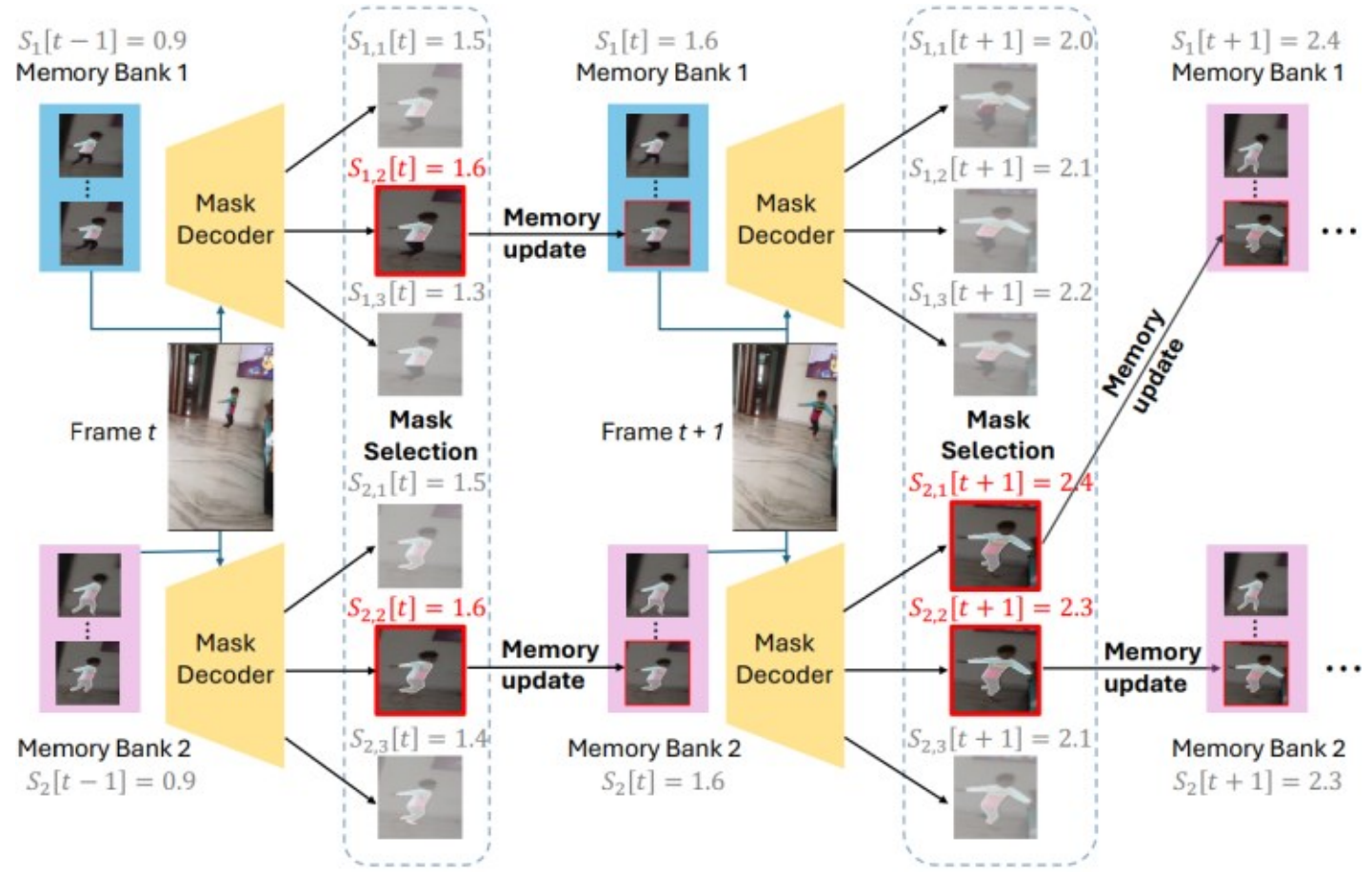

*Figure 15*. *The pipeline of constrained memory. The model incorporates multiple memory pathways, each with its own memory bank. A mask selection module selects masks based on occlusion status and cumulative IoU scores. The memory entries with the highest scores are then extracted to update the corresponding memory banks. Source: reproduced from [84].*

In [82-84], groups of previous frame features are entirely pruned or replaced. In contrast, MoSAM [85] introduces a spatial-temporal memory selection strategy that updates memory not only at the temporal level—as done in prior methods—but also at the spatial level by retaining only the most reliable regions. Specifically, at the temporal level, adjacent past frames are sampled at regular intervals, and remaining frame features are further filtered using IoU scores and occlusion scores computed from SAM2 for each frame. At the spatial level, probability segmentation maps from previous predictions are leveraged to discard potentially inaccurate regions, enabling the model to concentrate on more confident foreground areas.

## 4. Present: how to learn discriminative features for the current frame

Training a large model from scratch, or transfer learning a whole large model, are both time-intensive and costly. In addition, it requires curating a large-scale dataset to prevent overfitting during training. To address these challenges, many studies have explored parameter-efficient transfer learning (PETL) techniques to fine-tune pretrained foundation models such as SAM and SAM2. These methods aim to incorporate domain-specific knowledge while preserving the generalization capabilities of the original models by introducing lightweight, low-parameter modules that adapt existing features without modifying the full model. Currently, two widely adopted PETL techniques are Adapters [86] and Low-Rank Adaptation (LoRA) [87], particularly in the context of SAM-based models for medical image segmentation (see **Figure 16**). The Adapter method introduces small bottleneck networks—called Adapters—into the transformer blocks of the image encoder in SAM or SAM2. LoRA, on the other hand, inserts parallel low-rank bottleneck modules alongside the transformer blocks of the image encoder.

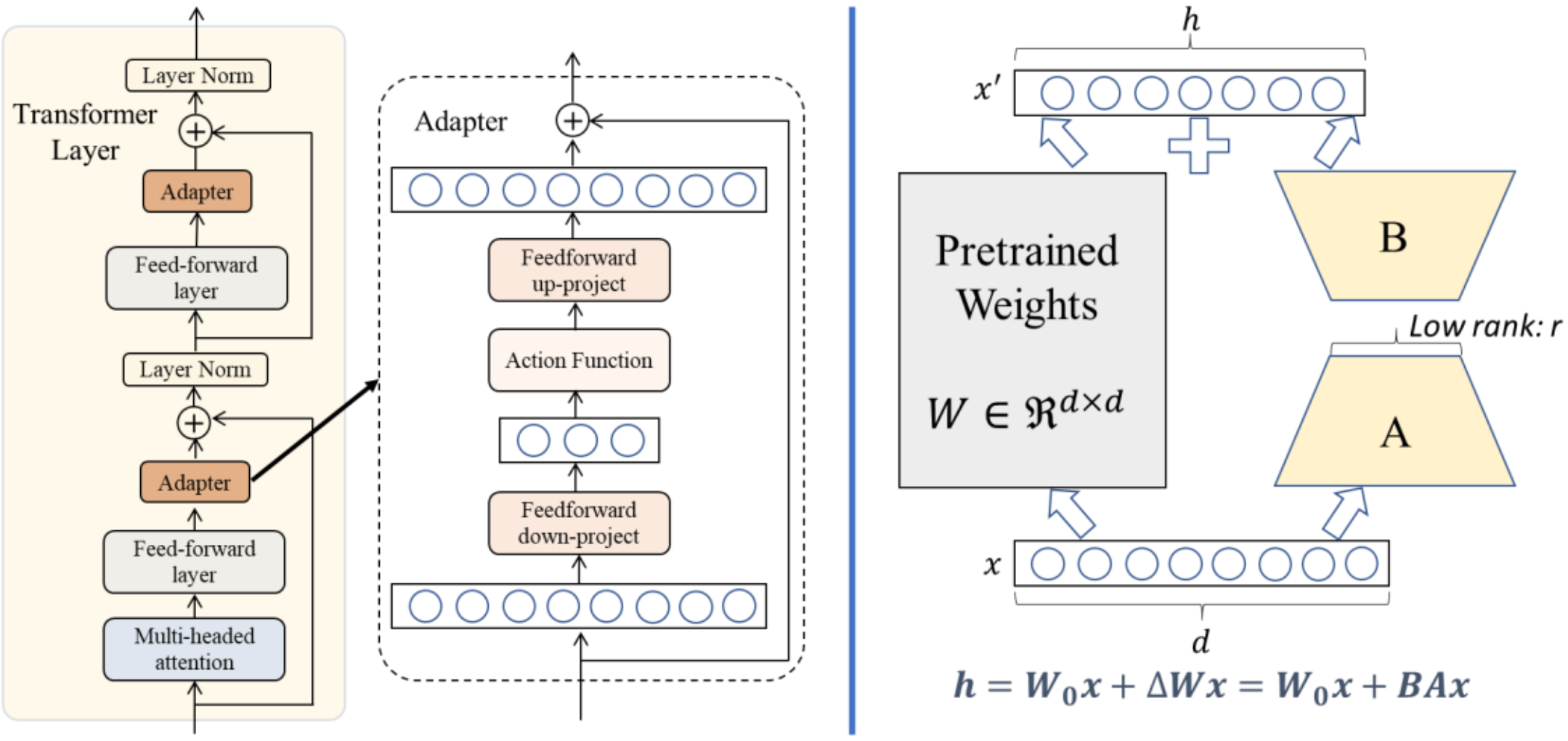

***Figure 16***. *Illustration of Adapter and LoRA. The adapters are inserted within the transformer blocks to enable lightweight fine-tuning. In contrast, LoRA introduces low-rank bottleneck layers in parallel with the original pre-trained sub-modules, typically within the transformer blocks.*

In the following section, we review various Adapter and LoRA variants used to fine-tune SAM/SAM2, aiming to provide a systematic and comprehensive overview of the progress in this area and to inspire future research toward optimal fine-tuning strategies for SAM2 in VOST tasks.

*(1) Adapter-based finetuning for SAM/SAM2*

The standard Adapter architecture typically comprises two multilayer perceptrons (MLPs) separated by a non-linear activation function. The first MLP reduces the channel dimensionality, while the second restores it to the original size (see **Figure 17** (a)). This bottleneck structure has been effectively utilized in the Medical SAM Adapter (Med-SA) [88], where Adapters are inserted into both the image encoder and prompt decoder to integrate domain-specific knowledge from medical imaging into the original SAM framework. Similarly, SAM-Adapter [89] and CWSAM [90] employ this Adapter design to address underperforming scenarios, such as camouflaged or shadowed objects and segmentation tasks in the synthetic aperture radar (SAR) domain, respectively.

To enhance local and channel-wise representation, SAM-Med2D [91] incorporates a channel-wise attention mechanism into a convolution-based bottleneck Adapter, enabling more effective feature adaptation during SAM fine-tuning (see **Figure 17** (b)). MA-SAM [92] and 3DSAM-Adapter [93] further advance this concept by integrating 3D convolutions and 3D depth-wise convolutions, respectively, within the Adapter modules to capture richer contextual information from volumetric medical imaging data (see **Figure 17** (c) and **Figure 17** (d)).

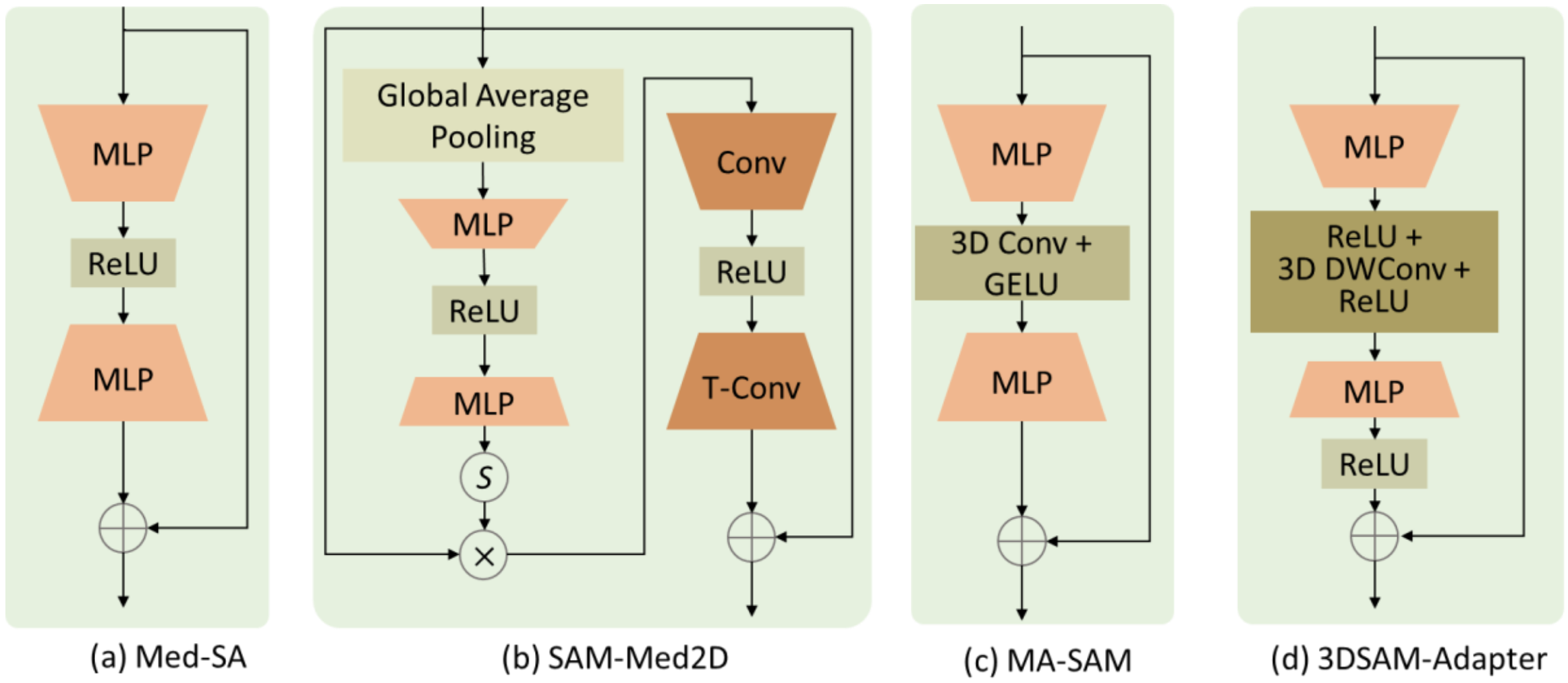

*Figure 17. Four types of adapter variants. (a) Standard bottleneck structure; (b) bottleneck with channel attention; (c) bottleneck with 3D convolution; and (d) bottleneck with 3D depthwise convolution. These variants are adopted in Med-SA, SAM-Med2D, MA-SAM, and 3DSAM-Adapter, respectively.*

In addition, several Adapter variants have been proposed that incorporate multi-scale feature integration [94], cross-branch architectural designs [95], and residual connections [96], among other innovations. While these Adapter-based approaches have demonstrated promising results in fine-tuning SAM for static image or volumetric segmentation tasks, their effectiveness within the SAM2 framework for video object segmentation and tracking (VOST) remains underexplored [97]. Given the temporal nature of VOST, further investigation is warranted—particularly regarding how to effectively incorporate historical information to guide the fine-tuning of features in the current frame.

*(2) LoRA-based fine-tuning for SAM/SAM2*

Low-Rank Adaptation (LoRA) is another widely adopted technique for fine-tuning large models. In the context of SAM, LoRA has been effectively employed in methods such as SAM-SP [98], SAMed [99], SonarSAM [100], and MediViSTA [101], where low-rank modules are inserted in parallel with each Transformer block within the image encoder (See **Figure 18**).

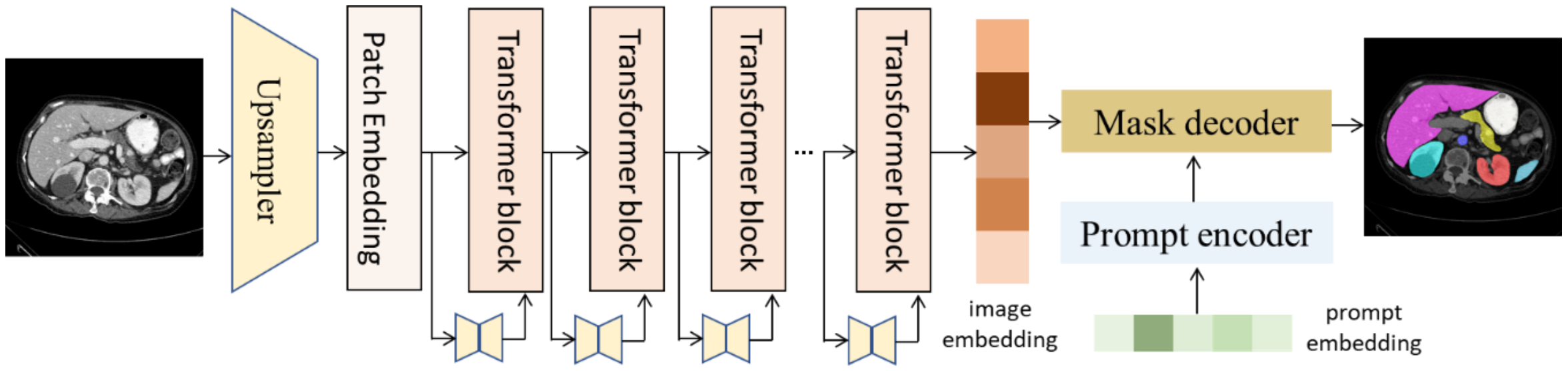

*Figure 18. The LoRA layers are integrated into SAMed to fine-tune the image encoder of SAM. Source: adopted from[99].*

In contrast, BLO-SAM [102] applies LoRA layers to the mask decoder and prompt encoder of SAM to mitigate overfitting in semantic segmentation. This is achieved by updating two separate sets of learnable parameters, each trained on distinct subsets of the dataset. These LoRA-based fine-tuning strategies have shown significant performance gains across various domain-specific segmentation tasks, underscoring their potential for efficient model adaptation with minimal computational overhead. However, the optimal strategy for deploying LoRA within the SAM architecture—and the precise impact of these modules on final segmentation performance—remains an open question, warranting further investigation.

In addition, recent studies have begun to explore the use of LoRA within the SAM2 framework [103, 104]. For instance, MLE-SAM [103] introduces a Mixture of Low-Rank Adaptation Experts (MoE-LoRA) within the image encoder of SAM2 to handle different input visual modalities. This architecture enables modality-specific adaptation by training separate LoRA modules for each modality and employing a dynamic routing mechanism to effectively integrate features across modalities. In VesselSAM [105] (see **Figure 19**), an Atrous Spatial Pyramid Pooling (ASPP) module [15] is incorporated into the LoRA design—referred to as AtrousLoRA—to enhance the model's ability to capture multi-scale contextual information, specifically tailored for aortic vessel segmentation tasks.

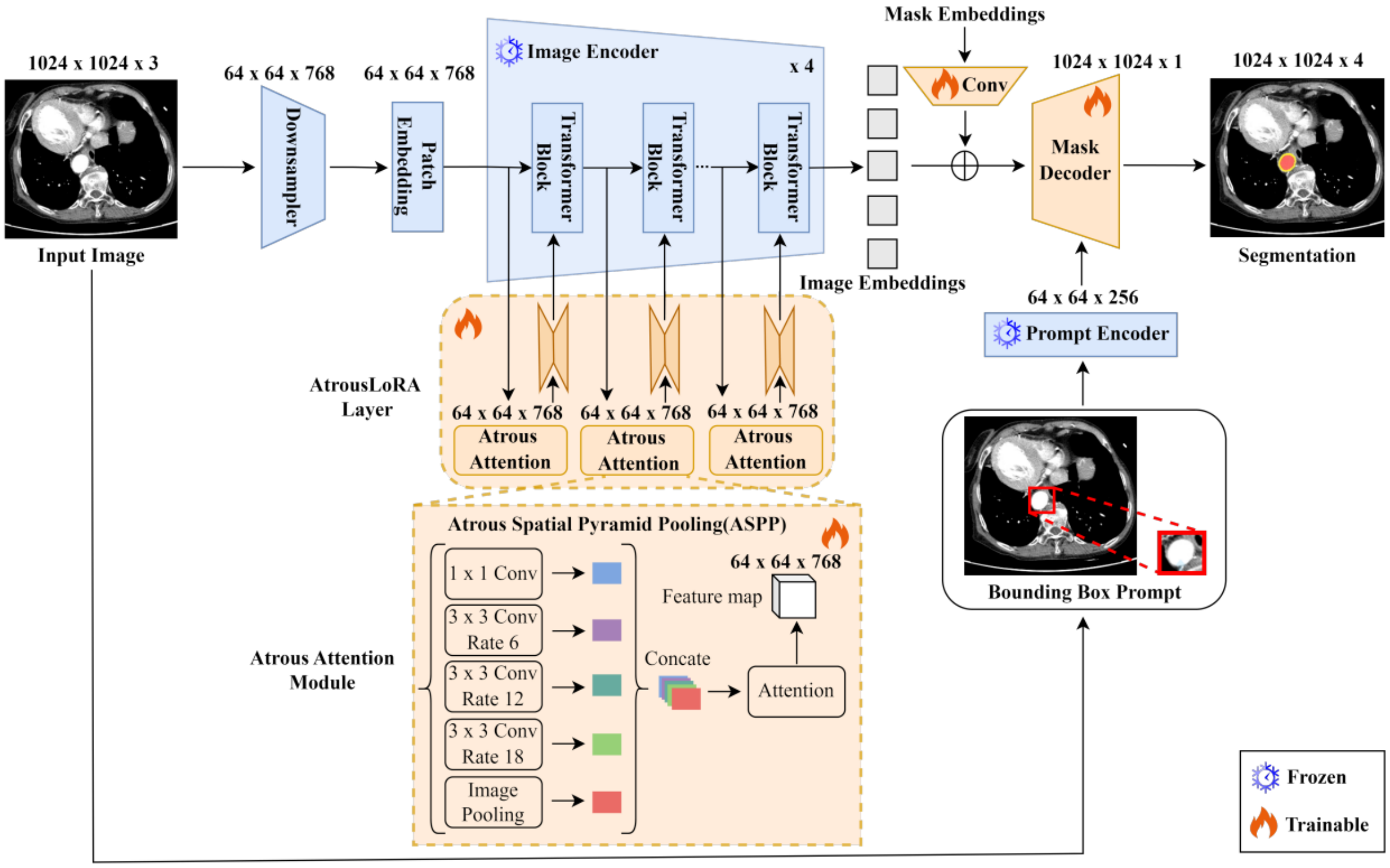

*Figure 19. The architecture of VesselSAM. AtrousLoRA modules are placed in parallel with transformer blocks, aiming to enhance multi-scale feature extraction through the Atrous Attention Module. Source: reproduced from[105].*

However, these approaches have primarily focused on static image segmentation. The application of LoRA-based fine-tuning in SAM2 for VOST remains underexplored. In particular, it is still unclear how the fine-tuned memory features influence the representation update of the current frame—a critical question for advancing LoRA's effectiveness in temporal settings like VOST.

## 5. Future: how to estimate the trajectory for the next frame

Motion tracking and trajectory estimation are fundamental components of object tracking, with classical methods such as optical flow and Kalman filtering widely used to ensure temporal consistency. These techniques have been incorporated into deep neural networks to improve video understanding tasks [55, 106-108]. More recently, transformer-based models such as PIPs++ [109], TAPIR [110], CoTracker [111], and CoTracker3 [112] have demonstrated strong performance in point tracking for dynamic scenes. However, as these approaches are tailored primarily for motion estimation rather than object segmentation, their applicability to VOST remains limited.

In the context of SAM-based segmentation, despite its strong generalization capability, SAM2 exhibits two key limitations when applied to VOST: (1) it can struggle to distinguish between multiple visually similar objects in crowded scenes, and (2) it may fail to capture fine-grained details of fast-moving objects. To address these challenges, several recent works have explored the integration of motion (trajectory) estimation into the SAM/SAM2 framework.

One representative example is SAM-PT [113], which enhances SAM's temporal awareness by propagating initial annotated points from the first frame—comprising both positive and negative samples—across the video sequence to produce object trajectories and occlusion scores (See **Figure 20**). These are then used as dynamic prompts for SAM in subsequent frames. The method involves four main steps: (1) generating query points from the first-frame annotation to indicate target and non-target regions; (2) employing point trackers to propagate these points and estimate trajectories along with occlusion scores; (3) using the resulting trajectories as prompts for SAM to segment each frame; and (4) optionally reinitializing point tracking based on predicted masks for better consistency in later frames.

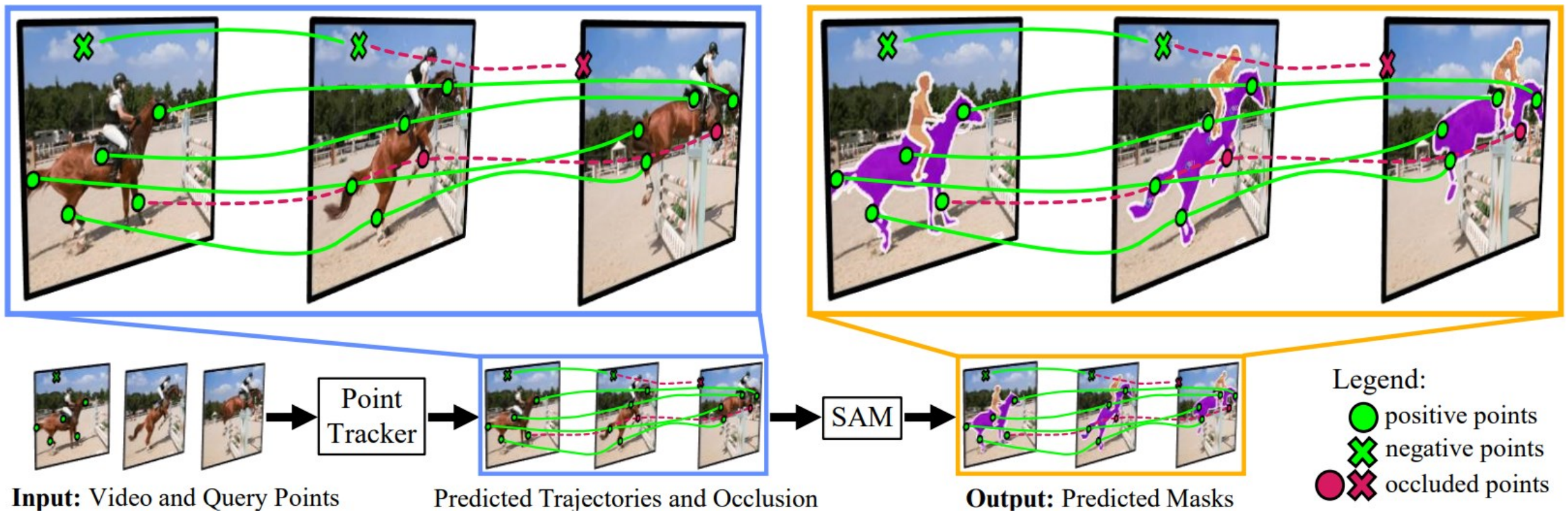

***Figure 20****. The pipeline of SAM-PT. The Point Tracker module is used to estimate positive points, negative points, and occluded points for guiding the segmentation and tracking process. Source: Reproduced from [113], with permission from IEEE.*

However, SAM-PT relies on externally pre-trained point trackers such as PIPS [114] or CoTracker, which must be integrated into the pipeline. This dependence on separate models hinders overall efficiency for fine-tuning and inference, as point tracking and segmentation are performed independently rather than in a unified framework.

In [115], a motion tracker is integrated with SAM2 to enable motion-aware object tracking and segmentation (see **Figure 21**). The framework begins by using two pre-trained models to generate 2D object tracks and corresponding depth maps. These points and depth maps are then processed through a motion encoder and track decoder, which refine the tracking points by filtering noise and decoupling motion and semantic information. The refined point prompts are fed into SAM2 to produce initial segmentation masks. Finally, dynamic trajectories belonging to the same object are grouped and reintroduced to SAM2, which refines the segmentation results and produces accurate masks for moving objects. However, similar to SAM-PT, this approach relies on an external motion tracker to obtain keypoint trajectories, which adds computational overhead and may limit efficiency.

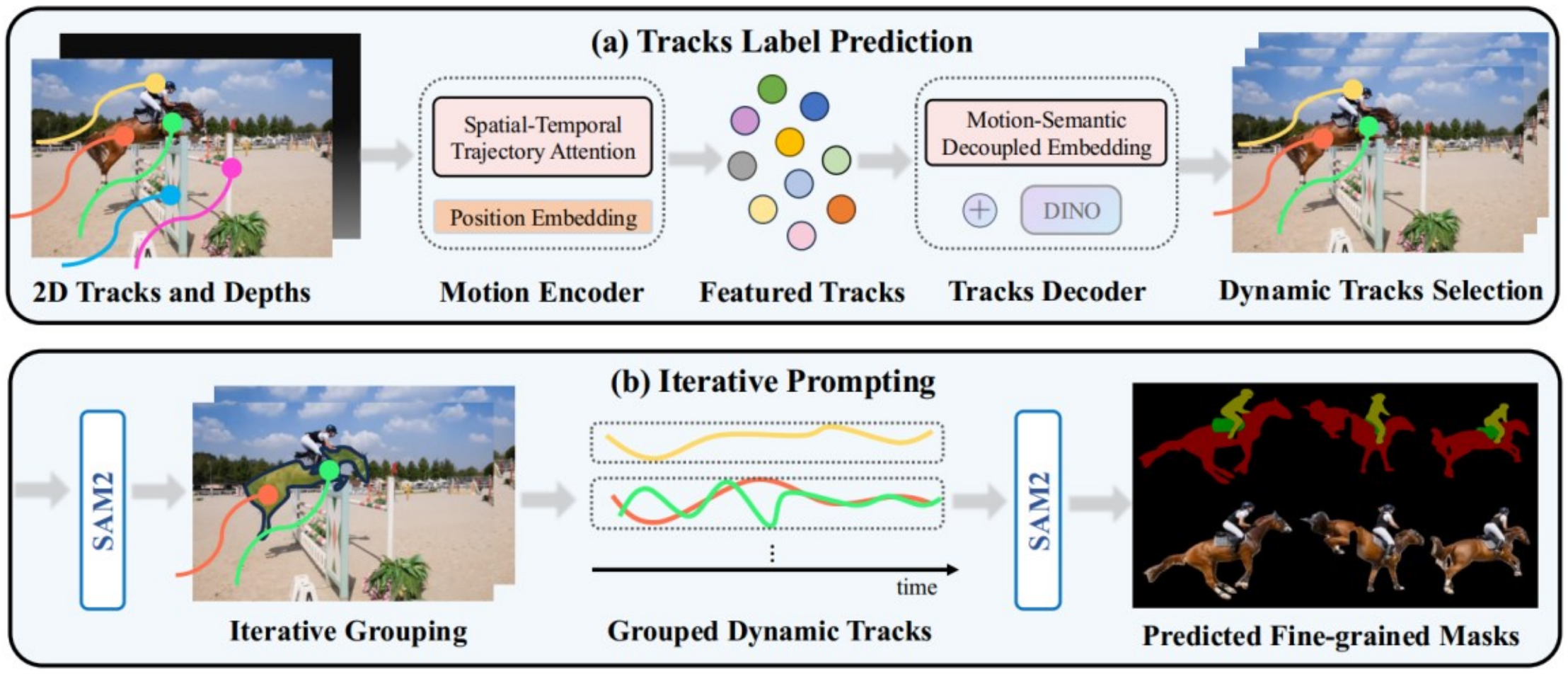

***Figure 21***. *Pipeline of Segment Any Motion in Videos. The method takes 2D object tracks and depth maps as input, encodes motion patterns, and extracts dynamic trajectories. SAM2 then groups these trajectories and generates fine-grained masks of moving objects. Source: Reproduced from [115], with permission from IEEE.*

To address the efficiency limitations of the above approach, SAMURAI [83] introduces two key enhancements. First, it incorporates temporal motion cues using a linear Kalman filter to improve predictions of bounding box positions and dimensions from predicted masks. Second, it employs an enhanced motion-aware memory selection mechanism that reduces error propagation in crowded scenes while maintaining computational efficiency. Together, these components significantly improve the model's capability to handle complex scenarios in moving object tracking without compromising speed.

In summary, these methods focus on the integration of motion dynamics to achieve accurate and efficient video object segmentation and tracking. While they have demonstrated effectiveness in certain scenarios, identifying an optimal and universally accurate approach for object tracking remains an open research question.

## 6. Related datasets and metrics for VOST

In this section, we briefly review several representative video segmentation datasets from both natural and medical scenarios. These datasets can serve as valuable resources for training and evaluating models for VOST. We also provide definitions of the related evaluation metrics.

Table 1 summarizes key public datasets commonly used for VOST in natural scenes. These datasets cover a wide range of challenging conditions, including occlusions, dense object interactions, long video sequences, and significant scale variations. For each dataset, we provide detailed information, including the total number of annotated objects, the number of videos, the total number of frames, and a brief description. Collectively, these datasets offer a comprehensive benchmark for assessing the performance and generalizability of VOST algorithms across diverse real-world scenarios.

**Table 1**. Summary of representative public datasets for VOST in natural scenes, including the number of annotated objects, videos, frames, and a brief description of each dataset.

| Dataset | Objects | Videos | Frames | Description |
|---|---|---|---|---|
| SegTrack | 6 | 6 | - | Early benchmark for object segmentation and tracking in short video clips |
| SegTrack-v2 | 24 | 14 | 976 | Expanded SegTrack with more objects and challenging sequences for segmentation and tracking |
| DAVIS16 | 50 | 50 | 3,455 | High-quality video object segmentation dataset, with challenges such as occlusion, fast motion, blur, and appearance change |
| DAVIS17 | 376 | 150 | 10,459 | Multi-object extension of DAVIS16 with complex scenes and multiple interacting objects |
| LVOS-V1 | 282 | 220 | 126,280 | Long video object segmentation benchmark focused on extended temporal consistency |
| LVOS-V2 | 1132 | 720 | 296,401 | Larger and more diverse long video segmentation dataset for evaluating scalability and robustness |
| YouTube-VOS | 7755 | 4453 | 120,532 | Large-scale dataset for video object segmentation, with diverse objects and real-world scenarios |
| MOSE | 5,200 | 2,149 | - | Complex video object segmentation with crowded scenes, severe occlusions, and dense object interactions |
| SA-V | - | 50.9K | 4.2M | Massive dataset covering multiple scenes and fine-grained details for segmentation at scale |

We summarize representative video datasets in medical scenarios in Table 2. The table provides an overview of key medical video datasets commonly used for segmentation and tracking tasks, highlighting their imaging modalities, annotated targets, and application domains. These datasets cover a range of tasks such as surgical tool segmentation, anatomical structure delineation, tumor tracking, polyp segmentation, and cell tracking, offering valuable benchmarks for the development and evaluation of medical VOST algorithms.

**Table 2**. Overview of widely used medical video datasets for segmentation and tracking tasks. The table details their imaging modality, annotated objects or targets, number of videos and frames, mean frames per second (mFPS), and a concise description of each dataset's focus and challenges.

| Dataset | Modality | Objects/Targets | video | Frames | mFPS | Description |
|---|---|---|---|---|---|---|
| Endovis17 | Endoscopy | Surgical instruments | 8 | 2,040 | 2 | Robotic instrument segmentation, to segment different articulated parts of a da Vinci robotic instrument |
| Endovis18 | Endoscopy | Surgical instruments | 90 | 10,700 | 2 | Robotic scene segmentation, including robotic instruments as well as anatomical objects and non-robotic surgical instruments |
| CAMUS | Ultrasound | Left ventricle, myocardium, atria | 500 | - | - | 2D apical two-chamber and apical four-chamber view video |
| EchoNet-Dynamic | Ultrasound | Left ventricle | 10,030 | - | 51 | 2D apical two-chamber view videos, only labeling end-systolic and end-diastolic phases |
| TrackRAD2025 | Cine-MRI | Tumor | 108 | - | 1~8 | Tumor tracking during radiotherapy |
| Endoscapes | Laparoscopy | Surgical tools, anatomy | 201 | 11,090 | 1 | Surgical scene segmentation, object detection, and critical view of safety assessment |
| SUN-SEG | Colonoscopy | Polyp | 1013 | 158,690 | 30 | Polyp segmentation |
| Polyp-Gen | Colonoscopy | Polyps | 2,225 | 8,037 | - | Polyp segmentation and tracking in GI endoscopy |
| Cell Tracking Challenge datasets | Microscopy | Individual cells | 52 | - | 25 | Cell segmentation and tracking in biological videos |

To assess segmentation and tracking performance, we adopt several standard metrics that measure both region-level accuracy and boundary precision as follows:

*(1) Intersection-over-Union (IoU or Jaccard Index)*

The IoU quantifies the overlap between the predicted segmentation mask and the 'ground-truth' mask. It is defined as:

$$J = \frac{|P \cap G|}{|P \cup G|} \tag{1}$$

Where $P$ is the set of predicted positive pixels and $G$ is the set of 'ground-truth' positive pixels. This metric provides a direct measure of how accurately the predicted regions align with the true regions.

*(2) Boundary F1 Score (F)*

The *F* score evaluates the precision and recall of predicted boundaries compared to 'ground-truth' boundaries. Specifically, it calculates the harmonic mean of boundary precision and boundary recall, offering insight into the model's ability to delineate fine contours:

$$F = \frac{2 \cdot Precision \cdot Recall}{Precision + Recall} \quad (2)$$

where precision and recall are computed on the boundary pixels.

*(3) J&F Score*

The J&F score is the arithmetic mean of the IoU (*J)* and Boundary F1 (*F*) scores:

$$J\&F = \frac{J + F}{2} \quad (3)$$

This composite score provides a balanced evaluation of both region overlap and boundary accuracy.

*(4) Dice Coefficient (Dice)*

The Dice coefficient measures the similarity between the predicted and 'ground-truth' masks, placing a greater emphasis on overlap. It is computed as:

$$Dice = \frac{2|P \cap G|}{|P| + |G|} \quad (4)$$

The Dice score is closely related to IoU but tends to be more sensitive to the overlap of smaller structures.

*(5) Challenge IoU (CIoU)*

CIoU follows the EndoVis2018 Challenge protocol and evaluates segmentation performance at the object level across the entire video. Instead of averaging frame-wise IoUs, CIoU computes the intersection-over-union over the entire spatiotemporal extent of each object. Specifically:

$$CIoU = \frac{|\bigcup_{i=1}^{N} P_i \cap \bigcup_{i=1}^{N} G_i|}{|\bigcup_{i=1}^{N} P_i \cup \bigcup_{i=1}^{N} G_i|} \quad (5)$$

where $N$ is the total number of frames, and $P_i$, $G_i$ denote the predicted and 'ground-truth' masks in frame $i$, respectively. This metric accumulates the predicted and 'ground-truth' regions over all frames for each object and then computes a global IoU, ensuring a temporally consistent, object-centric evaluation of segmentation quality.

(6) Success Rate (for bounding boxes)

Precision measures the Intersection over Union (IoU) between predicted and ground truth bounding boxes across a range of thresholds. It is typically reported as the proportion of frames in which the IoU exceeds a predefined threshold (e.g., 0.5):

$$Success\ Rate = \frac{Number\ of\ successful\ frames}{Total\ number\ of\ frames} \tag{6}$$

(7) Precision

The precision measures the center error distance between predicted and ground truth boxes. The proportion of frames where the center distance between predicted and ground truth bounding boxes is within a certain threshold (e.g., 20 pixels). The definition is as follows:

$$Precision = \frac{Number\ of\ frames\ with\ center\ error < threshold}{Total\ number\ of\ frames} \tag{7}$$

(6) Frames Per Second (FPS):

FPS quantifies the speed at which the model processes video frames during inference. This metric is critical for real-time applications, particularly in surgical settings where timely feedback is essential.

## 7. Discussion: Trends, Challenges, and Future Perspectives

(1) Memory extraction and updating

Memory plays a critical role in VOST, as historical information is essential for updating current frame features to support accurate object recognition and segmentation. Existing methods, such as LSTM [116], GRU [117], and Transformer-based models [118], typically extract memory features from images and/or prompts using various encoders to build memory banks. In particular, SAM2-based approaches tend to preserve historical information from a fixed set of frames, leveraging cross-attention to condition current features on past context. However, these methods often accumulate redundant or irrelevant information from previous frames, increasing computational overhead and potentially degrading performance.

Recent works such as SurgSAM2 and SAMURAI address this by pruning frame features based on similarity or motion scores, reducing the memory bank load and improving efficiency. Nonetheless, these strategies largely ignore temporal consistency and focus on fixed time windows, limiting their effectiveness for long-term VOST. Future directions may involve organizing memory into hierarchical components—such as sensory, short-term, and long-term memory—within the SAM2 framework to better support long video sequences. Such a design could enhance the model's ability to balance efficiency and temporal consistency over varying timescales. Furthermore, updating memory through residual learning could refine segmentation accuracy by incrementally correcting errors and emphasizing critical object variations [19], thereby reducing redundancy and focusing computation on informative changes [119].

In the context of medical imaging, effective memory extraction is highly dependent on the specific anatomical structures under investigation. For example, cardiac motion is predominantly periodic and involuntary, making it relatively stable for modeling. In contrast, respiratory dynamics involve a broader range of anatomical components—including the lungs, chest wall, hemi-diaphragms, upper airway, and adjacent abdominal organs—and are subject to voluntary modulation of breathing frequency and depth. Accurately capturing respiratory function remains a significant challenge in dynamic imaging, particularly in distinguishing normal tidal breathing from sporadic shallow or deep breaths. Joint kinematics introduces yet another complexity, with movement patterns governed by external stimuli and voluntary motion. These variations across anatomical systems suggest that task-specific prior knowledge can play a crucial role in guiding memory extraction and updating strategies for VOST in medical applications, improving both robustness and interpretability.

(2) Language-vision for multi-modality fusion in medical VOST

Combining the strengths of large language models (LLMs) with SAM/SAM2 for deeper interaction and multimodal fusion presents a promising direction for achieving more general and robust VOST solutions [120]. Several recent studies have begun exploring this integration. For example, RefSAM [121] employs a text-to-text model (T5) [122] to generate linguistic embeddings as prompts for SAM. SAM-Track [42] leverages Grounding-DINO [123], an open-set object detector capable of fusing language and vision modalities, to produce bounding box prompts for SAM. Similar approaches, such as SAMWISE [124] and SAM2-LOVE [125], aim to incorporate language cues to improve contextual understanding and enhance

VOST performance. However, to the best of our knowledge, no comparable work exists for medical VOST—that is, bridging language and vision for medical video object segmentation and tracking. A key barrier is the lack of large-scale, publicly available video-and-language datasets in the medical domain. In the future, we hope the creation of such datasets and the development of multimodal medical VOST models will unlock new opportunities for intelligent and generalizable medical video analysis.

(3) Prior knowledge for improving motion estimation

Prior knowledge can provide valuable motion cues that enhance contextual understanding of object dynamics in video sequences. In SAM-based methods for VOST, several works have focused on leveraging prior knowledge to support automatic prompt generation for subsequent frames. These approaches generally operate under the assumption that the positions of tracked objects do not undergo abrupt changes between consecutive frames. Representative examples include SAM-PD [67], PA-SAM [126], and DAM4SAM[127], where motion continuity is utilized to guide prompt placement and refinement.

In SAM2-based methods for VOST, prior knowledge is typically incorporated at a deeper level, not only to guide prompt generation but also to inform memory updating, temporal attention, and motion consistency modeling. By explicitly encoding assumptions about smooth object trajectories and temporal coherence, these methods can achieve more robust segmentation and tracking, particularly in challenging scenarios involving occlusions, deformation, or appearance changes [128].

Moreover, hybrid intelligence—the integration of natural and artificial intelligence—is emerging in medical imaging through the use of eye-tracking technologies in contexts where it is intuitively applicable. Eye-tracking patterns can serve not only as prompt signals but also as a window into cognitive processes, offering valuable natural-intelligence priors that enhance object detection and segmentation[129, 130]. Future directions may further explore integrating learned priors from large-scale video datasets or incorporating physics-based constraints to improve motion estimation within the SAM2 framework. Such advancements could further enhance the adaptability and precision of SAM2-based VOST systems in real-world medical applications.

(4) Efficient object tracking and segmentation

Real-time performance is a critical requirement for VOST, as it reflects both inference speed and computational efficiency [131]. Prior efforts to enhance efficiency have explored lightweight image encoders through knowledge distillation, as seen in MobileSAM [132], EfficientSAM [133], and SAM-Lightening [134], or by adopting compact architectures like RepViT-SAM [135] and SqueezeSAM [136].

However, these approaches primarily target 2D static image segmentation. While SAM2 introduces a streamlined transformer design with memory modules to accelerate video segmentation, the heavy image encoder still poses challenges for deployment in memory-constrained or resource-limited environments [137]. Future work may benefit from exploring complementary techniques such as pruning, quantization, and model refactorization [138], to further enhance efficiency without compromising segmentation and tracking accuracy in VOST.

(5) Practice-oriented medical benchmarks and datasets for VOST

Clinically grounded benchmarks are critical resources for medical video object tracking and segmentation, as they underpin rigorous evaluation, model generalization, and safe clinical use [139]. General-purpose vision benchmarks do not capture the unique complexities of clinical tasks, so models trained on generic data often struggle to generalize to real-world healthcare scenarios [140, 141].

To address this gap, benchmark tasks should span the spectrum of medical tasks and modalities, with clear delineation across planar video, volumetric tomographic data, and 2D–3D registration pipelines for image-guided interventions [141].

For VOST on 2D video modalities, tasks could involve using X-ray fluoroscopy, ultrasound, and endoscopy, dedicated datasets would contend with modality-specific noise such as continuous organ deformation and inherently low-contrast anatomical boundaries, challenges compounded by sparse annotations due to the labor-intensive labeling of medical videos.

Similarly, VOST on 3D imaging modalities such as CT and MRI require benchmarks for segmenting and tracking structures across volumetric data or time, where algorithms face large data volumes, subtle tissue contrast, and limited labeled examples [142]. Moreover, VOST using 2D-on-3D alignment, such as motion-compensated 2D fluoroscopy registered to a 3D scan, introduces cross-modal alignment issues and out-of-plane motion that no current general benchmark adequately evaluates [141, 143]. In addition, work in real-time radiotherapy systems, as exemplified by CyberKnife Xsight Lung Tracking [144, 145] in stereotactic radiotherapy, demonstrates the feasibility of markerless projection-driven tumor tracking. These domain-specific challenges highlight why introducing tailored medical video segmentation and tracking benchmarks is crucial for robust model development. By expanding coverage to include realistic noise, deformation, low-contrast conditions, and limited-label scenarios, such benchmarks enable more rigorous evaluation of algorithm performance and ensure models learn to handle the variability of clinical environments, strengthening their generalization. Ultimately, they are essential for clinical translation;

high-impact reports emphasize that well-designed medical AI benchmarks can drive higher solution quality and patient-benefiting adoption of AI in healthcare [139]. In short, future diverse and specialized VOST benchmarks reflecting actual medical conditions will help state-of-the-art models success in practice, underscoring the need for new medical video benchmarks to guide generalizable and safe clinical AI deployment.

(6) End-to-end segmentation and tracking joint learning

Future work can move from decoupled medical pipelines to unified training that jointly optimizes spatial segmentation and temporal association. Treating the tasks separately forfeits cross-task cues such as appearance consistency, motion priors, and shape constraints, which can regularize both objectives and reduce drift. Recent systems such as BioMedParse [146] exemplify end-to-end pipelines that integrate detection, segmentation, and classification in biomedical images, illustrating the feasibility of end-to-end learning. Complementary evidence from end-to-end video instance segmentation VisTR [118] and cell-level joint tracking models Cell-TRACTR [147] shows that transformer architectures can directly learn sequence-level correspondences without post-processing, reinforcing the feasibility of unified objectives for medical sequences.

(7) Detection-assisted screening and pipeline automation

Integrating detection into VOST pipelines can replace manual prompts with automatic detection such as Grounding-DINO [123] ,lowering annotation cost and enabling population-scale screening. A detection-first cascade localizes candidate lesions or instruments in long 2D videos and 3D series; these proposals then condition promptable segmenters and temporal trackers to yield precise masks and stable trajectories. Detection-assisted screening and pipeline automation will introduce a robust initialization mechanism for video object segmentation and tracking, enabling trackers to recover automatically from drift or occlusion and maintain accurate object identity over long sequences. Moreover, by providing reliable region proposals and re-grounding segmentation at key frames, detection integration will further improve VOST performance and resilience in complex clinical videos.

## 8. Conclusion

In this review, we examined the development of video object segmentation and tracking (VOST) based on the Segment Anything Model (SAM) and its successor, SAM2. We structured our analysis around three temporal dimensions: (1) past—how memory representations are stored and updated; (2) present—how current frame features are extracted and optimized; and (3) future—how motion and trajectory prediction inform upcoming segmentations. While recent SAM-based approaches have made significant strides in generalization, temporal coherence, and inference speed, several challenges remain. We discussed these limitations and outlined promising research directions for advancing prompt-driven segmentation in videos. We hope this survey serves as a valuable reference and inspiration for future work in this rapidly evolving field.

**Acknowledgements**

The study was supported by the US National Institutes of Health (R01 CA240808, R01 CA258987, R01 EB034691, and R01 CA280135).

**Reference**

1 Krizhevsky, A., Sutskever, I., and Hinton, G.E.: 'Imagenet classification with deep convolutional neural networks', Advances in neural information processing systems, 2012, 25

2 He, K., Zhang, X., Ren, S., and Sun, J.: 'Deep residual learning for image recognition', in Editor (Ed.)^(Eds.): 'Book Deep residual learning for image recognition' (2016, edn.), pp. 770-778

3 Vaswani, A., Shazeer, N., Parmar, N., Uszkoreit, J., Jones, L., Gomez, A.N., Kaiser, Ł., and Polosukhin, I.: 'Attention is all you need', Advances in neural information processing systems, 2017, 30

4 Simonyan, K., and Zisserman, A.: 'Very deep convolutional networks for large-scale image recognition', in Editor (Ed.)^(Eds.): 'Book Very deep convolutional networks for large-scale image recognition' (Computational and Biological Learning Society, 2015, edn.), pp.

5 Huang, G., Liu, Z., Van Der Maaten, L., and Weinberger, K.Q.: 'Densely connected convolutional networks', in Editor (Ed.)^(Eds.): 'Book Densely connected convolutional networks' (2017, edn.), pp. 4700-4708

6 Hu, J., Shen, L., and Sun, G.: 'Squeeze-and-excitation networks', in Editor (Ed.)^(Eds.): 'Book Squeeze-and-excitation networks' (2018, edn.), pp. 7132-7141

7 Ren, S., He, K., Girshick, R., and Sun, J.: 'Faster R-CNN: Towards real-time object detection with region proposal networks', IEEE transactions on pattern analysis and machine intelligence, 2016, 39, (6), pp. 1137-1149

8 Redmon, J., Divvala, S., Girshick, R., and Farhadi, A.: 'You only look once: Unified, real-time object detection', in Editor (Ed.)^(Eds.): 'Book You only look once: Unified, real-time object detection' (2016, edn.), pp. 779-788

9 Carion, N., Massa, F., Synnaeve, G., Usunier, N., Kirillov, A., and Zagoruyko, S.: 'End-to-end object detection with transformers', in Editor (Ed.)^(Eds.): 'Book End-to-end object detection with transformers' (Springer, 2020, edn.), pp. 213-229

10 He, K., Gkioxari, G., Dollár, P., and Girshick, R.: 'Mask r-cnn', in Editor (Ed.)^(Eds.): 'Book Mask r-cnn' (2017, edn.), pp. 2961-2969

11 Udupa, J.K., Liu, T., Jin, C., Zhao, L., Odhner, D., Tong, Y., Agrawal, V., Pednekar, G., Nag, S., and Kotia, T.: 'Combining natural and artificial intelligence for robust automatic anatomy segmentation: Application in neck and thorax auto‐contouring', Med Phys, 2022, 49, (11), pp. 7118-7149

12 Zheng, L., Zhang, H., Sun, S., Chandraker, M., Yang, Y., and Tian, Q.: 'Person re-identification in the wild', in Editor (Ed.)^(Eds.): 'Book Person re-identification in the wild' (2017, edn.), pp. 1367-1376

13 Ronneberger, O., Fischer, P., and Brox, T.: 'U-net: Convolutional networks for biomedical image segmentation', in Editor (Ed.)^(Eds.): 'Book U-net: Convolutional networks for biomedical image segmentation' (Springer, 2015, edn.), pp. 234-241

14 Long, J., Shelhamer, E., and Darrell, T.: 'Fully convolutional networks for semantic segmentation', in Editor (Ed.)^(Eds.): 'Book Fully convolutional networks for semantic segmentation' (2015, edn.), pp. 3431-3440

15 Chen, L.-C., Papandreou, G., Kokkinos, I., Murphy, K., and Yuille, A.L.: 'Deeplab: Semantic image segmentation with deep convolutional nets, atrous convolution, and fully connected crfs', IEEE transactions on pattern analysis and machine intelligence, 2017, 40, (4), pp. 834-848

16 Chen, J., Mei, J., Li, X., Lu, Y., Yu, Q., Wei, Q., Luo, X., Xie, Y., Adeli, E., and Wang, Y.: 'TransUNet: Rethinking the U-Net architecture design for medical image segmentation through the lens of transformers', Medical Image Analysis, 2024, 97, pp. 103280

17 LeCun, Y., Bengio, Y., and Hinton, G.: 'Deep learning', nature, 2015, 521, (7553), pp. 436-444

18 Schmidhuber, J.: 'Deep learning in neural networks: An overview', Neural networks, 2015, 61, pp. 85-117

19 Xu, G., Wang, X., Wu, X., Leng, X., and Xu, Y.: 'Development of residual learning in deep neural networks for computer vision: A survey', Engineering Applications of Artificial Intelligence, 2025, 142, pp. 109890

20 Karpathy, A., Toderici, G., Shetty, S., Leung, T., Sukthankar, R., and Fei-Fei, L.: 'Large-scale video classification with convolutional neural networks', in Editor (Ed.)^(Eds.): 'Book Large-scale video classification with convolutional neural networks' (2014, edn.), pp. 1725-1732

21 Kay, W., Carreira, J., Simonyan, K., Zhang, B., Hillier, C., Vijayanarasimhan, S., Viola, F., Green, T., Back, T., and Natsev, P.: 'The kinetics human action video dataset', arXiv preprint arXiv:1705.06950, 2017

22 Feichtenhofer, C., Fan, H., Malik, J., and He, K.: 'Slowfast networks for video recognition', in Editor (Ed.)^(Eds.): 'Book Slowfast networks for video recognition' (2019, edn.), pp. 6202-6211

23 Cheng, H.K., and Schwing, A.G.: 'Xmem: Long-term video object segmentation with an atkinson-shiffrin memory model', in Editor (Ed.)^(Eds.): 'Book Xmem: Long-term video object segmentation with an atkinson-shiffrin memory model' (Springer, 2022, edn.), pp. 640-658

24 Liu, M., Zhu, M., White, M., Li, Y., and Kalenichenko, D.: 'Looking fast and slow: Memory-guided mobile video object detection', arXiv preprint arXiv:1903.10172, 2019

25 Li, P., and Jin, J.: 'Time3d: End-to-end joint monocular 3d object detection and tracking for autonomous driving', in Editor (Ed.)^(Eds.): 'Book Time3d: End-to-end joint monocular 3d object detection and tracking for autonomous driving' (2022, edn.), pp. 3885-3894

26 Guo, Y., Wang, H., Hu, Q., Liu, H., Liu, L., and Bennamoun, M.: 'Deep learning for 3d point clouds: A survey', IEEE transactions on pattern analysis and machine intelligence, 2020, 43, (12), pp. 4338-4364

27 Milletari, F., Navab, N., and Ahmadi, S.-A.: 'V-net: Fully convolutional neural networks for volumetric medical image segmentation', in Editor (Ed.)^(Eds.): 'Book V-net: Fully convolutional neural networks for volumetric medical image segmentation' (Ieee, 2016, edn.), pp. 565-571

28 Hatamizadeh, A., Tang, Y., Nath, V., Yang, D., Myronenko, A., Landman, B., Roth, H.R., and Xu, D.: 'Unetr: Transformers for 3d medical image segmentation', in Editor (Ed.)^(Eds.): 'Book Unetr: Transformers for 3d medical image segmentation' (2022, edn.), pp. 574-584

29 Patil, P.W., Dudhane, A., Kulkarni, A., Murala, S., Gonde, A.B., and Gupta, S.: 'An unified recurrent video object segmentation framework for various surveillance environments', IEEE Transactions on Image

Processing, 2021, 30, pp. 7889-7902
30 Gao, M., Zheng, F., Yu, J.J., Shan, C., Ding, G., and Han, J.: 'Deep learning for video object segmentation: a review', Artificial Intelligence Review, 2023, 56, (1), pp. 457-531
31 Zhu, J., Qi, Y., and Wu, J.: 'Medical sam 2: Segment medical images as video via segment anything model 2', arXiv preprint arXiv:2408.00874, 2024
32 Yao, R., Lin, G., Xia, S., Zhao, J., and Zhou, Y.: 'Video object segmentation and tracking: A survey', ACM Transactions on Intelligent Systems and Technology (TIST), 2020, 11, (4), pp. 1-47
33 Oh, S.W., Lee, J.-Y., Xu, N., and Kim, S.J.: 'Video object segmentation using space-time memory networks', in Editor (Ed.)^(Eds.): 'Book Video object segmentation using space-time memory networks' (2019, edn.), pp. 9226-9235
34 Cheng, H.K., Tai, Y.-W., and Tang, C.-K.: 'Rethinking space-time networks with improved memory coverage for efficient video object segmentation', Advances in neural information processing systems, 2021, 34, pp. 11781-11794
35 Seong, H., Oh, S.W., Lee, J.-Y., Lee, S., Lee, S., and Kim, E.: 'Hierarchical memory matching network for video object segmentation', in Editor (Ed.)^(Eds.): 'Book Hierarchical memory matching network for video object segmentation' (2021, edn.), pp. 12889-12898
36 Lu, X., Wang, W., Danelljan, M., Zhou, T., Shen, J., and Van Gool, L.: 'Video object segmentation with episodic graph memory networks', in Editor (Ed.)^(Eds.): 'Book Video object segmentation with episodic graph memory networks' (Springer, 2020, edn.), pp. 661-679
37 Brown, T., Mann, B., Ryder, N., Subbiah, M., Kaplan, J.D., Dhariwal, P., Neelakantan, A., Shyam, P., Sastry, G., and Askell, A.: 'Language models are few-shot learners', Advances in neural information processing systems, 2020, 33, pp. 1877-1901
38 Ouyang, L., Wu, J., Jiang, X., Almeida, D., Wainwright, C., Mishkin, P., Zhang, C., Agarwal, S., Slama, K., and Ray, A.: 'Training language models to follow instructions with human feedback', Advances in neural information processing systems, 2022, 35, pp. 27730-27744
39 Solaiman, I., Brundage, M., Clark, J., Askell, A., Herbert-Voss, A., Wu, J., Radford, A., Krueger, G., Kim, J.W., and Kreps, S.: 'Release strategies and the social impacts of language models', arXiv preprint arXiv:1908.09203, 2019
40 Kirillov, A., Mintun, E., Ravi, N., Mao, H., Rolland, C., Gustafson, L., Xiao, T., Whitehead, S., Berg, A.C., and Lo, W.-Y.: 'Segment anything', in Editor (Ed.)^(Eds.): 'Book Segment anything' (2023, edn.), pp. 4015-4026
41 Yang, J., Gao, M., Li, Z., Gao, S., Wang, F., and Zheng, F.: 'Track anything: Segment anything meets videos', arXiv preprint arXiv:2304.11968, 2023
42 Cheng, Y., Li, L., Xu, Y., Li, X., Yang, Z., Wang, W., and Yang, Y.: 'Segment and track anything', arXiv preprint arXiv:2305.06558, 2023
43 Rajič, F., Ke, L., Tai, Y.W., Tang, C.K., Danelljan, M., and Yu, F.: 'Segment anything meets point tracking', in Editor (Ed.)^(Eds.): 'Book Segment anything meets point tracking' (IEEE, 2025, edn.), pp. 9302-9311
44 Zhu, J., Chen, Z., Hao, Z., Chang, S., Zhang, L., Wang, D., Lu, H., Luo, B., He, J.-Y., and Lan, J.-P.: 'Tracking anything in high quality', arXiv preprint arXiv:2307.13974, 2023
45 Ravi, N., Gabeur, V., Hu, Y.-T., Hu, R., Ryali, C., Ma, T., Khedr, H., Rädle, R., Rolland, C., and Gustafson, L.: 'Sam 2: Segment anything in images and videos', arXiv preprint arXiv:2408.00714, 2024
46 Zhou, T., Porikli, F., Crandall, D.J., Van Gool, L., and Wang, W.: 'A survey on deep learning technique for video segmentation', IEEE transactions on pattern analysis and machine intelligence, 2022, 45, (6), pp. 7099-7122
47 Hou, B., Liu, Y., Ling, N., Ren, Y., and Liu, L.: 'A survey of efficient deep learning models for moving object segmentation', APSIPA Transactions on Signal and Information Processing, 2023, 12, (1)
48 Zhou, Y., Sun, G., Li, Y., Benini, L., and Konukoglu, E.: 'When sam2 meets video camouflaged object

segmentation: A comprehensive evaluation and adaptation', arXiv preprint arXiv:2409.18653, 2024
49 Zhang, Y., and Shen, Z.: 'Unleashing the potential of sam2 for biomedical images and videos: A survey', arXiv preprint arXiv:2408.12889, 2024
50 Jiaxing, Z., and Hao, T.: 'SAM2 for Image and Video Segmentation: A Comprehensive Survey', arXiv preprint arXiv:2503.12781, 2025
51 Zhang, C., Cui, Y., Lin, W., Huang, G., Rong, Y., Liu, L., and Shan, S.: 'Segment anything for videos: A systematic survey', arXiv preprint arXiv:2408.08315, 2024
52 He, K., Chen, X., Xie, S., Li, Y., Dollár, P., and Girshick, R.: 'Masked autoencoders are scalable vision learners', in Editor (Ed.)^(Eds.): 'Book Masked autoencoders are scalable vision learners' (2022, edn.), pp. 16000-16009
53 Ryali, C., Hu, Y.-T., Bolya, D., Wei, C., Fan, H., Huang, P.-Y., Aggarwal, V., Chowdhury, A., Poursaeed, O., and Hoffman, J.: 'Hiera: A hierarchical vision transformer without the bells-and-whistles', in Editor (Ed.)^(Eds.): 'Book Hiera: A hierarchical vision transformer without the bells-and-whistles' (PMLR, 2023, edn.), pp. 29441-29454
54 Wu, J., and Xu, M.: 'One-prompt to segment all medical images', in Editor (Ed.)^(Eds.): 'Book One-prompt to segment all medical images' (2024, edn.), pp. 11302-11312
55 Perazzi, F., Khoreva, A., Benenson, R., Schiele, B., and Sorkine-Hornung, A.: 'Learning video object segmentation from static images', in Editor (Ed.)^(Eds.): 'Book Learning video object segmentation from static images' (2017, edn.), pp. 2663-2672
56 Revaud, J., Weinzaepfel, P., Harchaoui, Z., and Schmid, C.: 'Epicflow: Edge-preserving interpolation of correspondences for optical flow', in Editor (Ed.)^(Eds.): 'Book Epicflow: Edge-preserving interpolation of correspondences for optical flow' (2015, edn.), pp. 1164-1172
57 Bailer, C., Taetz, B., and Stricker, D.: 'Flow fields: Dense correspondence fields for highly accurate large displacement optical flow estimation', in Editor (Ed.)^(Eds.): 'Book Flow fields: Dense correspondence fields for highly accurate large displacement optical flow estimation' (2015, edn.), pp. 4015-4023
58 Maninis, K.-K., Pont-Tuset, J., Arbeláez, P., and Van Gool, L.: 'Convolutional oriented boundaries', in Editor (Ed.)^(Eds.): 'Book Convolutional oriented boundaries' (Springer, 2016, edn.), pp. 580-596
59 Yang, Z., Wei, Y., and Yang, Y.: 'Associating objects with transformers for video object segmentation', Advances in Neural Information Processing Systems, 2021, 34, pp. 2491-2502
60 Yang, Z., Miao, J., Wei, Y., Wang, W., Wang, X., and Yang, Y.: 'Scalable video object segmentation with identification mechanism', IEEE Transactions on Pattern Analysis and Machine Intelligence, 2024
61 Hu, P., Wang, G., Kong, X., Kuen, J., and Tan, Y.-P.: 'Motion-guided cascaded refinement network for video object segmentation', in Editor (Ed.)^(Eds.): 'Book Motion-guided cascaded refinement network for video object segmentation' (2018, edn.), pp. 1400-1409
62 Li, X., and Loy, C.C.: 'Video object segmentation with joint re-identification and attention-aware mask propagation', in Editor (Ed.)^(Eds.): 'Book Video object segmentation with joint re-identification and attention-aware mask propagation' (2018, edn.), pp. 90-105
63 Tokmakov, P., Alahari, K., and Schmid, C.: 'Learning video object segmentation with visual memory', in Editor (Ed.)^(Eds.): 'Book Learning video object segmentation with visual memory' (2017, edn.), pp. 4481-4490
64 Han, J., Yang, L., Zhang, D., Chang, X., and Liang, X.: 'Reinforcement cutting-agent learning for video object segmentation', in Editor (Ed.)^(Eds.): 'Book Reinforcement cutting-agent learning for video object segmentation' (2018, edn.), pp. 9080-9089
65 Zhang, H., Li, F., Liu, S., Zhang, L., Su, H., Zhu, J., Ni, L.M., and Shum, H.-Y.: 'Dino: Detr with improved denoising anchor boxes for end-to-end object detection', arXiv preprint arXiv:2203.03605, 2022
66 Xie, J., Yang, C., Xie, W., and Zisserman, A.: 'Moving object segmentation: All you need is sam (and flow)', in Editor (Ed.)^(Eds.): 'Book Moving object segmentation: All you need is sam (and flow)' (2024,

edn.), pp. 162-178
67 Zhou, T., Luo, W., Ye, Q., Shi, Z., and Chen, J.: ‘Sam-pd: How far can sam take us in tracking and segmenting anything in videos by prompt denoising’, arXiv preprint arXiv:2403.04194, 2024
68 Cheng, H.K., Oh, S.W., Price, B., Schwing, A., and Lee, J.-Y.: ‘Tracking anything with decoupled video segmentation’, in Editor (Ed.)^(Eds.): ‘Book Tracking anything with decoupled video segmentation’ (2023, edn.), pp. 1316-1326
69 Mansoori, M., Shahabodini, S., Abouei, J., Plataniotis, K.N., and Mohammadi, A.: ‘Self-Prompting Polyp Segmentation in Colonoscopy Using Hybrid YOLO-SAM2 Model’, in Editor (Ed.)^(Eds.): ‘Book Self-Prompting Polyp Segmentation in Colonoscopy Using Hybrid YOLO-SAM2 Model’ (IEEE, 2025, edn.), pp. 1-5
70 Pandey, S., Chen, K.-F., and Dam, E.B.: ‘Comprehensive multimodal segmentation in medical imaging: Combining yolov8 with sam and hq-sam models’, in Editor (Ed.)^(Eds.): ‘Book Comprehensive multimodal segmentation in medical imaging: Combining yolov8 with sam and hq-sam models’ (2023, edn.), pp. 2592-2598
71 Atkinson, R.C., and Shiffrin, R.M.: ‘Human memory: A proposed system and its control processes’: ‘Psychology of learning and motivation’ (Elsevier, 1968), pp. 89-195
72 Ke, L., Ye, M., Danelljan, M., Tai, Y.-W., Tang, C.-K., and Yu, F.: ‘Segment anything in high quality’, Advances in Neural Information Processing Systems, 2023, 36, pp. 29914-29934
73 Deng, X., Wu, H., Zeng, R., and Qin, J.: ‘MemSAM: taming segment anything model for echocardiography video segmentation’, in Editor (Ed.)^(Eds.): ‘Book MemSAM: taming segment anything model for echocardiography video segmentation’ (2024, edn.), pp. 9622-9631
74 Yang, Z., and Yang, Y.: ‘Decoupling features in hierarchical propagation for video object segmentation’, Advances in Neural Information Processing Systems, 2022, 35, pp. 36324-36336
75 Zhou, J., Pang, Z., and Wang, Y.-X.: ‘Rmem: Restricted memory banks improve video object segmentation’, in Editor (Ed.)^(Eds.): ‘Book Rmem: Restricted memory banks improve video object segmentation’ (2024, edn.), pp. 18602-18611
76 Bekuzarov, M., Bermudez, A., Lee, J.-Y., and Li, H.: ‘Xmem++: Production-level video segmentation from few annotated frames’, in Editor (Ed.)^(Eds.): ‘Book Xmem++: Production-level video segmentation from few annotated frames’ (2023, edn.), pp. 635-644
77 Mei, H., Zhang, P., and Shou, M.Z.: ‘SAM-I2V: Upgrading SAM to Support Promptable Video Segmentation with Less than 0.2% Training Cost’, in Editor (Ed.)^(Eds.): ‘Book SAM-I2V: Upgrading SAM to Support Promptable Video Segmentation with Less than 0.2% Training Cost’ (2025, edn.), pp. 3417-3426
78 Li, M., Hu, L., Xiong, Z., Zhang, B., Pan, P., and Liu, D.: ‘Recurrent dynamic embedding for video object segmentation’, in Editor (Ed.)^(Eds.): ‘Book Recurrent dynamic embedding for video object segmentation’ (2022, edn.), pp. 1332-1341
79 Chen, Z., Zhang, L., Hu, P., Lu, H., and He, Y.: ‘MaskTrack: Auto-Labeling and Stable Tracking for Video Object Segmentation’, IEEE Transactions on Neural Networks and Learning Systems, 2024
80 Zhu, J., Hamdi, A., Qi, Y., Jin, Y., and Wu, J.: ‘Medical sam 2: Segment medical images as video via segment anything model 2’, arXiv preprint arXiv:2408.00874, 2024
81 Yan, Z., Sun, W., Zhou, R., Yuan, Z., Zhang, K., Li, Y., Liu, T., Li, Q., Li, X., and He, L.: ‘Biomedical sam 2: Segment anything in biomedical images and videos’, arXiv preprint arXiv:2408.03286, 2024
82 Liu, H., Zhang, E., Wu, J., Hong, M., and Jin, Y.: ‘Surgical sam 2: Real-time segment anything in surgical video by efficient frame pruning’, arXiv preprint arXiv:2408.07931, 2024
83 Yang, C.-Y., Huang, H.-W., Chai, W., Jiang, Z., and Hwang, J.-N.: ‘Samurai: Adapting segment anything model for zero-shot visual tracking with motion-aware memory’, arXiv preprint arXiv:2411.11922, 2024
84 Ding, S., Qian, R., Dong, X., Zhang, P., Zang, Y., Cao, Y., Guo, Y., Lin, D., and Wang, J.: ‘Sam2long: Enhancing sam 2 for long video segmentation with a training-free memory tree’, arXiv preprint

arXiv:2410.16268, 2024
85 Yang, Q., Yao, Y., Cui, M., and Bo, L.: 'MoSAM: Motion-Guided Segment Anything Model with Spatial-Temporal Memory Selection', arXiv preprint arXiv:2505.00739, 2025
86 Houlsby, N., Giurgiu, A., Jastrzebski, S., Morrone, B., De Laroussilhe, Q., Gesmundo, A., Attariyan, M., and Gelly, S.: 'Parameter-efficient transfer learning for NLP', in Editor (Ed.)^(Eds.): 'Book Parameter-efficient transfer learning for NLP' (PMLR, 2019, edn.), pp. 2790-2799
87 Hu, E.J., Shen, Y., Wallis, P., Allen-Zhu, Z., Li, Y., Wang, S., Wang, L., and Chen, W.: 'Lora: Low-rank adaptation of large language models', ICLR, 2022, 1, (2), pp. 3
88 Wu, J., Wang, Z., Hong, M., Ji, W., Fu, H., Xu, Y., Xu, M., and Jin, Y.: 'Medical sam adapter: Adapting segment anything model for medical image segmentation', Medical image analysis, 2025, 102, pp. 103547
89 Chen, T., Zhu, L., Deng, C., Cao, R., Wang, Y., Zhang, S., Li, Z., Sun, L., Zang, Y., and Mao, P.: 'Sam-adapter: Adapting segment anything in underperformed scenes', in Editor (Ed.)^(Eds.): 'Book Sam-adapter: Adapting segment anything in underperformed scenes' (2023, edn.), pp. 3367-3375
90 Pu, X., Jia, H., Zheng, L., Wang, F., and Xu, F.: 'ClassWise-SAM-adapter: Parameter efficient fine-tuning adapts segment anything to SAR domain for semantic segmentation', IEEE Journal of Selected Topics in Applied Earth Observations and Remote Sensing, 2025
91 Cheng, J., Ye, J., Deng, Z., Chen, J., Li, T., Wang, H., Su, Y., Huang, Z., Chen, J., and Jiang, L.: 'Sam-med2d', arXiv preprint arXiv:2308.16184, 2023
92 Chen, C., Miao, J., Wu, D., Zhong, A., Yan, Z., Kim, S., Hu, J., Liu, Z., Sun, L., and Li, X.: 'Ma-sam: Modality-agnostic sam adaptation for 3d medical image segmentation', Medical Image Analysis, 2024, 98, pp. 103310
93 Gong, S., Zhong, Y., Ma, W., Li, J., Wang, Z., Zhang, J., Heng, P.-A., and Dou, Q.: '3DSAM-adapter: Holistic adaptation of SAM from 2D to 3D for promptable tumor segmentation', Medical Image Analysis, 2024, 98, pp. 103324
94 Wu, Y., Wang, Z., Yang, X., Kang, H., He, A., and Li, T.: 'Trans-SAM: Transfer Segment Anything Model to medical image segmentation with Parameter-Efficient Fine-Tuning', Knowledge-Based Systems, 2025, 310, pp. 112909
95 Tu, Z., Gu, L., Wang, X., and Jiang, B.: 'Ultrasound sam adapter: Adapting sam for breast lesion segmentation in ultrasound images', arXiv preprint arXiv:2404.14837, 2024
96 Chen, J., Yu, X., Liu, S., Chen, T., Wang, W., Jeon, G., and He, B.: 'Tunnel SAM adapter: Adapting segment anything model for tunnel water leakage inspection', Geohazard Mechanics, 2024, 2, (1), pp. 29-36
97 Chen, T., Lu, A., Zhu, L., Ding, C., Yu, C., Ji, D., Li, Z., Sun, L., Mao, P., and Zang, Y.: 'Sam2-adapter: Evaluating & adapting segment anything 2 in downstream tasks: Camouflage, shadow, medical image segmentation, and more', arXiv preprint arXiv:2408.04579, 2024
98 Zhou, C., Ning, K., Shen, Q., Zhou, S., Yu, Z., and Wang, H.: 'Sam-sp: Self-prompting makes sam great again', arXiv preprint arXiv:2408.12364, 2024
99 Zhang, K., and Liu, D.: 'Customized segment anything model for medical image segmentation', arXiv preprint arXiv:2304.13785, 2023
100 Wang, L., Ye, X., Zhu, L., Wu, W., Zhang, J., Xing, H., and Hu, C.: 'When sam meets sonar images', IEEE Geoscience and Remote Sensing Letters, 2024
101 Kim, S., Jin, P., Chen, C., Kim, K., Lyu, Z., Ren, H., Kim, S., Liu, Z., Zhong, A., and Liu, T.: 'MediViSTA: Medical Video Segmentation via Temporal Fusion SAM Adaptation for Echocardiography', IEEE Journal of Biomedical and Health Informatics, 2025
102 Zhang, L., Liang, Y., Zhang, R., Javadi, A., and Xie, P.: 'BLO-SAM: bi-level optimization based finetuning of the segment anything model for overfitting-preventing semantic segmentation', in Editor (Ed.)^(Eds.): 'Book BLO-SAM: bi-level optimization based finetuning of the segment anything model for overfitting-preventing semantic segmentation' (2024, edn.), pp.

103 Zhu, C., Xiao, B., Shi, L., Xu, S., and Zheng, X.: 'Customize Segment Anything Model for Multi-Modal Semantic Segmentation with Mixture of LoRA Experts', arXiv preprint arXiv:2412.04220, 2024
104 Chen, X., Wang, C., Ning, H., Zhang, M., Shen, M., and Li, S.: 'Sam-octa2: Layer sequence octa segmentation with fine-tuned segment anything model 2', in Editor (Ed.)^(Eds.): 'Book Sam-octa2: Layer sequence octa segmentation with fine-tuned segment anything model 2' (IEEE, 2025, edn.), pp. 1-5
105 Iltaf, A., Ahmed, R.M., Zhang, Z., Li, B., and Zhou, S.: 'VesselSAM: Leveraging SAM for Aortic Vessel Segmentation with LoRA and Atrous Attention', arXiv preprint arXiv:2502.18185, 2025
106 Liu, J., and Guo, G.: 'Vehicle localization during GPS outages with extended Kalman filter and deep learning', IEEE Transactions on Instrumentation and Measurement, 2021, 70, pp. 1-10
107 de Araujo, P.R.M., Elhabiby, M., Givigi, S., and Noureldin, A.: 'A novel method for land vehicle positioning: Invariant kalman filters and deep-learning-based radar speed estimation', IEEE Transactions on Intelligent Vehicles, 2023, 8, (9), pp. 4275-4286
108 Zhai, M., Xiang, X., Lv, N., and Kong, X.: 'Optical flow and scene flow estimation: A survey', Pattern Recognition, 2021, 114, pp. 107861
109 Zheng, Y., Harley, A.W., Shen, B., Wetzstein, G., and Guibas, L.J.: 'Pointodyssey: A large-scale synthetic dataset for long-term point tracking', in Editor (Ed.)^(Eds.): 'Book Pointodyssey: A large-scale synthetic dataset for long-term point tracking' (2023, edn.), pp. 19855-19865
110 Doersch, C., Yang, Y., Vecerik, M., Gokay, D., Gupta, A., Aytar, Y., Carreira, J., and Zisserman, A.: 'Tapir: Tracking any point with per-frame initialization and temporal refinement', in Editor (Ed.)^(Eds.): 'Book Tapir: Tracking any point with per-frame initialization and temporal refinement' (2023, edn.), pp. 10061-10072
111 Karaev, N., Rocco, I., Graham, B., Neverova, N., Vedaldi, A., and Rupprecht, C.: 'Cotracker: It is better to track together', in Editor (Ed.)^(Eds.): 'Book Cotracker: It is better to track together' (Springer, 2024, edn.), pp. 18-35
112 Karaev, N., Makarov, I., Wang, J., Neverova, N., Vedaldi, A., and Rupprecht, C.: 'CoTracker3: Simpler and better point tracking by pseudo-labelling real videos', arXiv preprint arXiv:2410.11831, 2024
113 Rajic, F., Ke, L., Tai, Y.-W., Tang, C.-K., Danelljan, M., and Yu, F.: 'Segment anything meets point tracking', in Editor (Ed.)^(Eds.): 'Book Segment anything meets point tracking' (IEEE Computer Society, 2025, edn.), pp. 9302-9311
114 Harley, A.W., Fang, Z., and Fragkiadaki, K.: 'Particle video revisited: Tracking through occlusions using point trajectories', in Editor (Ed.)^(Eds.): 'Book Particle video revisited: Tracking through occlusions using point trajectories' (Springer, 2022, edn.), pp. 59-75
115 Huang, N., Zheng, W., Xu, C., Keutzer, K., Zhang, S., Kanazawa, A., and Wang, Q.: 'Segment Any Motion in Videos', in Editor (Ed.)^(Eds.): 'Book Segment Any Motion in Videos' (2025, edn.), pp. 3406-3416
116 Song, H., Wang, W., Zhao, S., Shen, J., and Lam, K.-M.: 'Pyramid dilated deeper convlstm for video salient object detection', in Editor (Ed.)^(Eds.): 'Book Pyramid dilated deeper convlstm for video salient object detection' (2018, edn.), pp. 715-731
117 Hong, L., Chen, W., Liu, Z., Zhang, W., Guo, P., Chen, Z., and Zhang, W.: 'Lvos: A benchmark for long-term video object segmentation', in Editor (Ed.)^(Eds.): 'Book Lvos: A benchmark for long-term video object segmentation' (2023, edn.), pp. 13480-13492
118 Wang, Y., Xu, Z., Wang, X., Shen, C., Cheng, B., Shen, H., and Xia, H.: 'End-to-end video instance segmentation with transformers', in Editor (Ed.)^(Eds.): 'Book End-to-end video instance segmentation with transformers' (2021, edn.), pp. 8741-8750
119 Liao, C., Zheng, X., Lyu, Y., Xue, H., Cao, Y., Wang, J., Yang, K., and Hu, X.: 'Memorysam: Memorize modalities and semantics with segment anything model 2 for multi-modal semantic segmentation', arXiv preprint arXiv:2503.06700, 2025
120 Yuan, H., Li, X., Zhang, T., Huang, Z., Xu, S., Ji, S., Tong, Y., Qi, L., Feng, J., and Yang, M.-H.: 'Sa2VA: Marrying SAM2 with LLaVA for Dense Grounded Understanding of Images and Videos', arXiv preprint

arXiv:2501.04001, 2025
121 Li, Y., Zhang, J., Teng, X., Lan, L., and Liu, X.: 'Refsam: Efficiently adapting segmenting anything model for referring video object segmentation', arXiv preprint arXiv:2307.00997, 2023
122 Raffel, C., Shazeer, N., Roberts, A., Lee, K., Narang, S., Matena, M., Zhou, Y., Li, W., and Liu, P.J.: 'Exploring the limits of transfer learning with a unified text-to-text transformer', Journal of machine learning research, 2020, 21, (140), pp. 1-67
123 Liu, S., Zeng, Z., Ren, T., Li, F., Zhang, H., Yang, J., Jiang, Q., Li, C., Yang, J., and Su, H.: 'Grounding dino: Marrying dino with grounded pre-training for open-set object detection', in Editor (Ed.)^(Eds.): 'Book Grounding dino: Marrying dino with grounded pre-training for open-set object detection' (Springer, 2024, edn.), pp. 38-55
124 Cuttano, C., Trivigno, G., Rosi, G., Masone, C., and Averta, G.: 'Samwise: Infusing wisdom in sam2 for text-driven video segmentation', in Editor (Ed.)^(Eds.): 'Book Samwise: Infusing wisdom in sam2 for text-driven video segmentation' (2025, edn.), pp. 3395-3405
125 Wang, Y., Xu, H., Liu, Y., Li, J., and Tang, Y.: 'SAM2-LOVE: Segment Anything Model 2 in Language-aided Audio-Visual Scenes', in Editor (Ed.)^(Eds.): 'Book SAM2-LOVE: Segment Anything Model 2 in Language-aided Audio-Visual Scenes' (2025, edn.), pp. 28932-28941
126 Xie, Z., Guan, B., Jiang, W., Yi, M., Ding, Y., Lu, H., and Zhang, L.: 'Pa-sam: Prompt adapter sam for high-quality image segmentation', in Editor (Ed.)^(Eds.): 'Book Pa-sam: Prompt adapter sam for high-quality image segmentation' (IEEE, 2024, edn.), pp. 1-6
127 Videnovic, J., Lukezic, A., and Kristan, M.: 'A distractor-aware memory for visual object tracking with sam2', in Editor (Ed.)^(Eds.): 'Book A distractor-aware memory for visual object tracking with sam2' (2025, edn.), pp. 24255-24264
128 Zhao, X., Pang, Y., Chang, S., Zhao, Y., Zhang, L., Lu, H., Fakhri, G.E., and Liu, X.: 'Inspiring the next generation of segment anything models: Comprehensively evaluate sam and sam 2 with diverse prompts towards context-dependent concepts under different scenes', arXiv preprint arXiv:2412.01240, 2024
129 Wang, B., Pan, H., Aboah, A., Zhang, Z., Keles, E., Torigian, D., Turkbey, B., Krupinski, E., Udupa, J., and Bagci, U.: 'Gazegnn: A gaze-guided graph neural network for chest x-ray classification', in Editor (Ed.)^(Eds.): 'Book Gazegnn: A gaze-guided graph neural network for chest x-ray classification' (2024, edn.), pp. 2194-2203
130 Wong, D., Wang, B., Durak, G., Tliba, M., Chaudhari, A., Chetouani, A., Cetin, A.E., Topel, C., Gennaro, N., and Vendrami, C.L.: 'Eyes Tell the Truth: GazeVal Highlights Shortcomings of Generative AI in Medical Imaging', in Editor (Ed.)^(Eds.): 'Book Eyes Tell the Truth: GazeVal Highlights Shortcomings of Generative AI in Medical Imaging' (2025, edn.), pp. 3167-3175
131 Chen, Y., Yildiz, Z., Li, Q., Chen, Y., Dong, H., Gu, H., Konz, N., and Mazurowski, M.A.: 'Accelerating Volumetric Medical Image Annotation via Short-Long Memory SAM 2', arXiv preprint arXiv:2505.01854, 2025
132 Zhang, C., Han, D., Qiao, Y., Kim, J.U., Bae, S.-H., Lee, S., and Hong, C.S.: 'Faster segment anything: Towards lightweight sam for mobile applications', arXiv preprint arXiv:2306.14289, 2023
133 Xiong, Y., Varadarajan, B., Wu, L., Xiang, X., Xiao, F., Zhu, C., Dai, X., Wang, D., Sun, F., and Iandola, F.: 'Efficientsam: Leveraged masked image pretraining for efficient segment anything', in Editor (Ed.)^(Eds.): 'Book Efficientsam: Leveraged masked image pretraining for efficient segment anything' (2024, edn.), pp. 16111-16121
134 Song, Y., Pu, B., Wang, P., Jiang, H., Dong, D., Cao, Y., and Shen, Y.: 'Sam-lightening: A lightweight segment anything model with dilated flash attention to achieve 30 times acceleration', arXiv preprint arXiv:2403.09195, 2024
135 Wang, A., Chen, H., Lin, Z., Han, J., and Ding, G.: 'Repvit-sam: Towards real-time segmenting anything', arXiv preprint arXiv:2312.05760, 2023
136 Varadarajan, B., Soran, B., Iandola, F., Xiang, X., Xiong, Y., Wu, L., Zhu, C., Krishnamoorthi, R., and

Chandra, V.: 'SqueezeSAM: User friendly mobile interactive segmentation', arXiv preprint arXiv:2312.06736, 2023
137 Zhou, C., Zhu, C., Xiong, Y., Suri, S., Xiao, F., Wu, L., Krishnamoorthi, R., Dai, B., Loy, C.C., and Chandra, V.: 'EdgeTAM: On-Device Track Anything Model', arXiv preprint arXiv:2501.07256, 2025
138 Sun, X., Liu, J., Shen, H.T., Zhu, X., and Hu, P.: 'On Efficient Variants of Segment Anything Model: A Survey', arXiv preprint arXiv:2410.04960, 2024
139 Yang, Y., Zhang, H., Gichoya, J.W., Katabi, D., and Ghassemi, M.: 'The limits of fair medical imaging AI in real-world generalization', Nature Medicine, 2024, 30, (10), pp. 2838-2848
140 Eche, T., Schwartz, L.H., Mokrane, F.-Z., and Dercle, L.: 'Toward generalizability in the deployment of artificial intelligence in radiology: role of computation stress testing to overcome underspecification', Radiology: Artificial Intelligence, 2021, 3, (6), pp. e210097
141 Markelj, P., Tomaževič, D., Likar, B., and Pernuš, F.: 'A review of 3D/2D registration methods for image-guided interventions', Medical image analysis, 2012, 16, (3), pp. 642-661
142 Singh, S.P., Wang, L., Gupta, S., Goli, H., Padmanabhan, P., and Gulyás, B.: '3D deep learning on medical images: a review', Sensors, 2020, 20, (18), pp. 5097
143 Gendrin, C., Furtado, H., Weber, C., Bloch, C., Figl, M., Pawiro, S.A., Bergmann, H., Stock, M., Fichtinger, G., and Georg, D.: 'Monitoring tumor motion by real time 2D/3D registration during radiotherapy', Radiotherapy and oncology, 2012, 102, (2), pp. 274-280
144 Klein, T.J., Gill, S., Ebert, M.A., Grogan, G., Smith, W., Alkhatib, Z., Geraghty, J., Scott, A.J., Brown, A., and Rowshanfarzad, P.: 'CyberKnife Xsight versus fiducial-based target-tracking: a novel 3D dosimetric comparison in a dynamic phantom', Radiation Oncology, 2022, 17, (1), pp. 154
145 Nakayama, M., Nishimura, H., Mayahara, H., Nakamura, M., Uehara, K., Tsudou, S., Harada, A., Akasaka, H., and Sasaki, R.: 'Clinical log data analysis for assessing the accuracy of the CyberKnife fiducial-free lung tumor tracking system', Practical radiation oncology, 2018, 8, (2), pp. e63-e70
146 Zhao, T., Gu, Y., Yang, J., Usuyama, N., Lee, H.H., Kiblawi, S., Naumann, T., Gao, J., Crabtree, A., and Abel, J.: 'A foundation model for joint segmentation, detection and recognition of biomedical objects across nine modalities', Nature methods, 2025, 22, (1), pp. 166-176
147 O'Connor, O.M., and Dunlop, M.J.: 'Cell-TRACTR: A transformer-based model for end-to-end segmentation and tracking of cells', PLOS Computational Biology, 2025, 21, (5), pp. e1013071